\documentclass[sigconf,nonacm]{acmart}
\AtBeginDocument{%
  \providecommand\BibTeX{{%
    \normalfont B\kern-0.5em{\scshape i\kern-0.25em b}\kern-0.8em\TeX}}}

\copyrightyear{2021}

\setcopyright{none}

\begin{document}

\title{Interpretable Additive Recurrent Neural Networks For Multivariate Clinical Time Series}

\author{Asif Rahman, Yale Chang, Jonathan Rubin}
\email{asif.rahman@philips.com, yale.chang@philips.com, jonathan.rubin@philips.com}
\affiliation{%
  \institution{Philips Research North America}
  \streetaddress{222 Jacobs St}
  \city{Cambridge}
  \state{MA}
  \country{USA}
  \postcode{02446}
}


\newcommand{\E}{\mathbb{E}}
\newcommand{\R}{\mathbb{R}}

\begin{abstract}
Time series models with recurrent neural networks (RNNs) can have high accuracy but are unfortunately difficult to interpret as a result of feature-interactions, temporal-interactions, and non-linear transformations. Interpretability is important in domains like healthcare where constructing models that provide insight into the relationships they have learned are required to validate and trust model predictions. We want accurate time series models that are interpretable where users can understand the contribution of individual input features. We present the Interpretable-RNN (I-RNN) that balances model complexity and accuracy by forcing the relationship between variables in the model to be additive. Interactions are restricted between hidden states of the RNN and additively combined at the final step. The I-RNN architecture specifically captures the unique characteristics of clinical time series, which are unevenly sampled in time, asynchronously acquired, and have missing data. Importantly, the hidden state activations represent feature coefficients that correlate with the prediction target and can be visualized as risk curves that capture the global relationship between individual input features and the outcome. We evaluate the I-RNN model on the Physionet 2012 Challenge dataset to predict in-hospital mortality, and on a real-world clinical decision support task: predicting hemodynamic interventions in the intensive care unit. I-RNN provides explanations in the form of global and local feature importances comparable to highly intelligible models like decision trees trained on hand-engineered features while significantly outperforming them. Additionally, I-RNN remains intelligible while providing accuracy comparable to state-of-the-art decay-based and interpolation-based recurrent time series models. The experimental results on real-world clinical datasets refute the myth that there is a tradeoff between accuracy and interpretability.
\end{abstract}



\maketitle

\section{Introduction}

Real world signals, like clinical time series, are irregularly sampled, multivariate, asynchronously acquired, have missing data, and are noisy. Multiple, often correlated signals, are necessary to fully predict a patient outcome but are observed separately and at different times. For example, in the medical domain, sequential measurements like nurse charted vital signs, laboratory measurements and interventions are not made at fixed intervals of time and measurements may be missing not at random (MNAR), thus reflecting patterns in clinical practice such that the missingness pattern itself may be informative.

In this context, recurrent neural networks (RNNs) have been widely used for clinical prediction \cite{liptonModelingMissingData2016,liptonLearningDiagnoseLSTM2017,choiDoctorAIPredicting2016,rajkomarScalableAccurateDeep2018,norgeotAssessmentDeepLearning2019,wangDevelopmentValidationDeep2019}. The predictions from RNN models, however, are difficult to explain because of nonlinear feature interactions and temporal interactions. The model complexity limits their use in exploratory data analysis and is a barrier for wider adoption in high-stakes decision making where model transparency and interpretability is critical. By interpretability we mean to understand the contribution of individual features to the predictions, also called local feature importance, as well as how features are associated with the outcome, also called global feature importance \cite{caruanaIntelligibleModelsHealthCare2015,louIntelligibleModelsClassification2012}. We want RNN models that have high accuracy and also explain their predictions in a way humans can understand.


\subsection{Related work}

Prior work on modeling time series with RNNs in the medical domain sometimes ignore the interpretability aspect in favor of high accuracy models, like in \cite{tomasevClinicallyApplicableApproach2019} and \cite{rajkomarScalableAccurateDeep2018}. Post-hoc methods like occlusion analysis are presented as tools to qualify how the black box model behaves under artificial conditions \cite{norgeotAssessmentDeepLearning2019}. These feature erasure approaches evaluate the direction of change in model performance after removing a feature. Another class of model explanations trains a second (post-hoc) model to explain the first black-box model (e.g. model distillation or mimic learning) \cite{tanLearningGlobalAdditive2018,ribeiroWhyShouldTrust2016,lundbergConsistentIndividualizedFeature2019,cheInterpretableDeepModels2017}.

A popular class of models focused on interpretability uses nonlinear attribution methods like attention mechanisms to provide instance-wise feature attribution scores \cite{tomasevClinicallyApplicableApproach2019,rajkomarScalableAccurateDeep2018,choiRETAINInterpretablePredictive2017,choiDoctorAIPredicting2016,shermanLeveragingClinicalTimeSeries2018,guoExploringInterpretableLSTM2019}. One way to instill interpretability is to model features (channels) independently. Modeling the dynamics in individual variables independently before combining them has been shown to increase model performance \cite{harutyunyanMultitaskLearningBenchmarking2019a,zhangModellingEHRTimeseries2019a}. Guo et. al. \cite{guoExploringInterpretableLSTM2019} introduced the IMV-LSTM model, which trained RNNs to learn variable-wise hidden states followed by an attention layer to weight the contribution of different time steps. The temporal-attention weighted hidden states were then non-linearly combined using feature-wise attention weights. Other notable attention-based RNN models include Patient2Vec and RAIM \cite{zhangPatient2VecPersonalizedInterpretable2018,xuRAIMRecurrentAttentive2018} that similarly provide attribution scores over time and features using attention weights. Choi et. al. \cite{choiRETAINInterpretablePredictive2017} introduced the RETAIN model, which uses attention to calculate the contribution of each feature at each time step. RETAIN is unique in that the coefficients of the output are correlated with the prediction target, which is different from IMV-LSTM that calculates the feature importance by the attention each variable receives. These attention-based explanations give local feature importances, however, attention-based feature attributions are influenced by feature interactions that can be difficult to disentangle \cite{rudinStopExplainingBlack2019}. Attention weights are not causal and can have weak correlation with the target, unlike coefficients of a logistic regression that are correlated with the prediction target.

For a general review of interpretability methods with neural networks, see the review by Chakraborty et. al.  \cite{chakrabortyInterpretabilityDeepLearning2017} as well as recently proposed methods for explaining risk scores in clinical prediction models by Hardt et. al. \cite{hardtExplainingIncreasePredicted2020a}, which also highlights some of the challenges (biases) in modeling clinical time series.

\subsection{Main contributions}

In this paper we present the Interpretable-RNN (I-RNN), that gives model explanations in the form of local and global feature importances over time. We achieve an intelligible model for multivariate time series by unifying prior concepts related to modeling irregularly sampled time series and interpretability into a single RNN architecture and introducing a new constraint on the shape of hidden states that further improves interpretability. Specifically, by 1) restricting interactions between features so that each RNN hidden state activation can be mapped to an input feature, 2) using information about how long a measurement has been missing to decay the contribution of expired features, 3) introducing a novel short-term memory process to control the shape and dynamics of the hidden state activations in-between observations, and finally 4) by additively combining the contribution of individual hidden states so the feature contributions can be mapped back to the inputs. In contrast to prior work on interpretable RNNs, I-RNN does not rely on attention mechanisms and does not need post-hoc attribution methods to interpret the predictions. Importantly, I-RNN hidden states, like the coefficients of a logistic regression, are directly correlated with the prediction target and our method yields directly interpretable feature-wise risk curves without needing post-hoc explanation methods. Our goal is to demonstrate intelligibility without sacrificing performance. Therefore, we devote much of this paper to analyzing the quality of the learned risk functions on a real-world clinical decision support task. We argue that complex RNNs are not necessarily more accurate and it is possible to achieve intelligible time series models with high predictive performance.

\section{Methods}

\subsection{Data processing}

\begin{figure}[t]
  \centering 
  \caption{Data processing to define the measurement vector $x_t$, the elapsed time vector $\delta_t$, and measurement indicator vector $m_t$. Missing and forward-filled values are in red.}
  \includegraphics[width=\columnwidth]{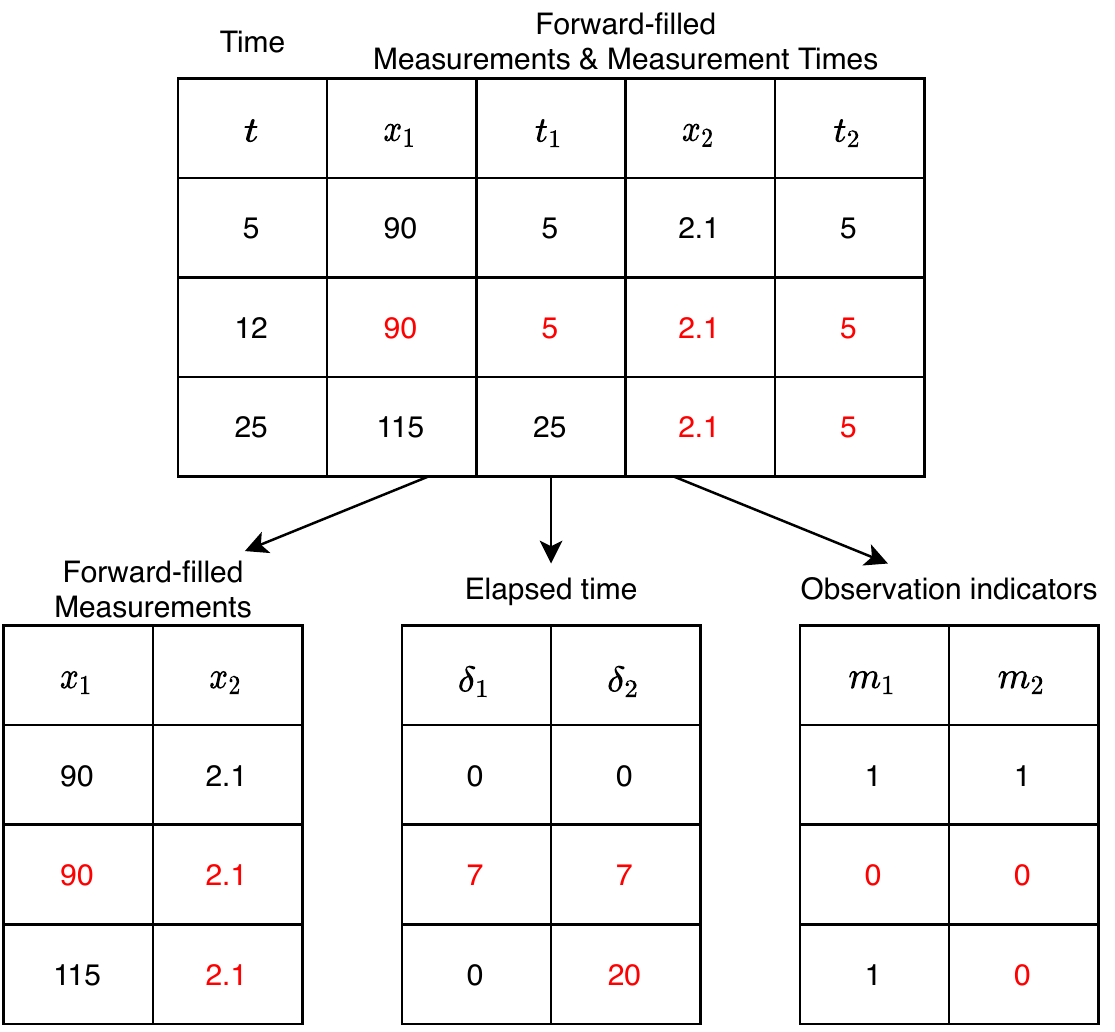}
  \label{fig:data}
\end{figure}

We address the general setting for time series classification/regression with missing values. Given a multivariate time series with $D$ variables of length $T$ as $X=(x_1,...,x_T)^T \in \mathbb{R}^{T \times D}$, where for each $t \in \{1,...,T\}$, $x_t \in \mathbb{R}^{D}$ represents observations at the $t$-th time step with $D$ features. $t$ are distinct observed time points of all variables. Values were transformed with log transformation if $|\text{skew}|>3.0$, z-score normalized, and clipped in the range $[-4, 4]$. Variables are forward filled indefinitely without expiration. Variables that were never observed are zero filled -- which is effectively mean filling since observations are z-score normalized. Samples have different signal lengths and were zero--padded to a maximum fixed length signal.

In addition to the sequence of observations, we maintain a sequence of elapsed time between observations for each variable indicating the duration since each feature at time $t$ was last observed ($\delta_t \in \R^{D}$). For example, $\delta_{t}^{d}=0$ when feature $d$ is observed at time $t$ and $\delta_{t}^{d}>0$ when a feature is forward filled. We use the elapsed time $\delta_t^d$ to inversely weight the contribution of the $d$-th feature over time. The elapsed time was normalized in the range $[0,1]$. Figure \ref{fig:data} illustrates how the measurement, elapsed time, and observation indicator matrices are created from a wide table containing both the value and time of measurements belonging to multiple variables.

The data format in Figure \ref{fig:data} addresses the unique characteristics of clinical time series. The elapsed time vector would allow irregularly sampled time series and asynchronous acquisition of different time series. Forward-filling, zero-filling and zero-padding are used to impute missing data.

\begin{figure}[t]
  \centering 
  \caption{Feature-specific risk scores are learned as independent channels then additively combined. A novel short-term memory process controls the shape of the risk scores such that a feature-specific risk is calculated as $h_t$ when an observation $x_t$ is made and then decays to a baseline $\mu_t$ at a rate $\gamma_t$ in-between observations.}
  \includegraphics[width=\columnwidth]{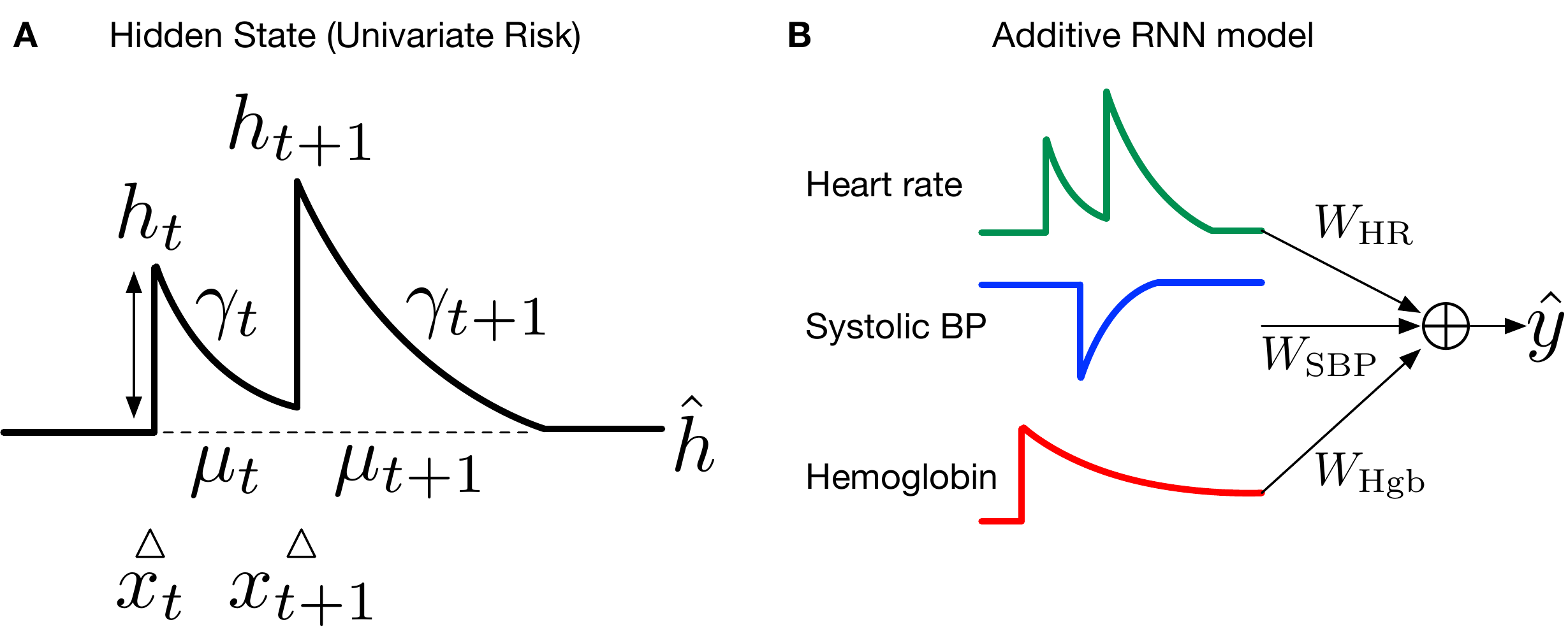}
  \label{fig:schematic} 
\end{figure}

\subsection{Model}

The I-RNN model considers each feature independently by restricting interactions between features and hidden states. First, we introduce a modification to the Gated Recurrent Unit (GRU) \cite{choLearningPhraseRepresentations2014}, called the MaskedGRU, that applies a mask $M$, which is an identity matrix, on the input-hidden ($W_{ir}, W_{iz}, W_{in} \in \R^{D \times D}$) and hidden-hidden ($W_{hr}, W_{hz}, W_{hn} \in \R^{D \times D}$) recurrent weight matrices. Constraining the recurrent weight matrices to be diagonal effectively makes the hidden states for each feature independent from the others. This mask allows us to control the model complexity by considering each input channel separately \cite{harutyunyanMultitaskLearningBenchmarking2019a,zhangModellingEHRTimeseries2019a}. In this paper, we use the GRU cell but the masking can be extended to other RNN formulations like long-short term memory (LSTM) cells.

\begin{equation}
r_t = \sigma((W_{ir} \odot M) x_t + (W_{hr} \odot M) \hat{h}_{t-1} + b_{r})
\end{equation}
\begin{equation}
z_t = \sigma((W_{iz} \odot M) x_t + (W_{hz} \odot M) \hat{h}_{t-1} + b_{z})
\end{equation}
\begin{equation}
n_t = \tanh((W_{in} \odot M) x_t + (W_{hn} \odot M) (r_t \odot \hat{h}_{t-1}) + b_{n})
\end{equation}
\begin{equation}
h_t = (1 - z_t) \odot \hat{h}_{t-1} + z_t \odot n_t
\end{equation}

The hidden state activation $h_t$ depends on past activations of the intensity function $\hat{h}_{t-1}$. The intensity function reaches an amplitude $h_t$ when a value is observed ($\delta_t=0$) and decays to a baseline $\mu_t$ at a rate $\gamma_t$ (Figure \ref{fig:schematic}). The baseline can be static ($\mu_t = 0$) or dynamic ($\mu_t \in \R^{D}$) when modeled as a function of the input $x_t$. The weights $W_\mu$ and $W_\gamma$ can also be constrained to be diagonal so there is no interaction between features when estimating the baseline or the decay rate.

\begin{equation}
	h_t=\text{MaskedGRU}(x_t, \hat{h}_{t-1})
\end{equation}
\begin{equation}
	\mu_t=W_{\mu}x_t + b_{\mu}
\end{equation}
\begin{equation}
	\gamma_t=\text{max}(0, W_\gamma\delta_t + b_\gamma)
\end{equation}
\begin{equation}
	\hat{h}_t=\mu_t + (h_t - \mu_t)\text{exp}(-\gamma_t)
\end{equation}

The final prediction is a weighted sum of the the intensity $\hat{h}_t \in R^{D}$ over each feature. This is the step where features are combined additively. For classification problems, $\hat{y}_t$ is in units of log-odds (before sigmoid transformation) and $W_{o}^{d}\hat{h}_{t}^{d}$ is the univariate contribution for feature $d$ at time $t$. The univariate contributions at each time step can be directly visualized as a local feature importance.

\begin{equation}
	\hat{y}_t=\sum_{d=1}^{D}{W_{o}^{d}\hat{h}_{t}^{d}} + b_o
\end{equation}

In practice, we use the prediction at the final time step $\hat{y}_T$ to calculate the loss function (binary cross entropy in our experiments), taking care to select the final time step since we have samples with variable signal length. 

\subsection{Global and Local Feature Importance}

How does I-RNN explain predictions? To explain how features relate to the global model output, we visualize the univariate contributions for each feature as $c_t^{d}=W_{o}^{d}\hat{h}_{t}^{d}+b_o$. Univariate risk curves representing the relationship between feature value and univariate contribution is calculated by averaging the feature contributions over all time steps ($u^d=\frac{1}{T}\sum_{t=1}^{T}{c_t^d}$) and plotting $(x^d, u^d)$ as in Figure \ref{fig:curves}. A ranked list of top features is visualized in Figure \ref{fig:global} by taking the mean of the absolute value of the univariate contributions over all patients: $\frac{1}{N}\sum_{i=1}^{N}{|u^d[n]|}$. Local feature contributions, which are sample specific, can be visualized by plotting the contributions for each feature over time $(t,c_t^d)$ as in Figure \ref{fig:example}.

\subsection{Baselines}

We compare the accuracy of the I-RNN model with a set of non-RNN baselines and state-of-the-art RNN models for irregularly sampled time series, including:
\begin{itemize}
	\item \textbf{Logistic} regression and gradient boosted decision trees (\textbf{XGBoost}) that are trained on hand engineered features summarizing the time series signals (maximum, minimum, mean, first value, last value, variance).
	\item \textbf{GRU-Forward}: GRU model using the forward-filled measurement vector $x_t$ as input.
	\item \textbf{GRU-Simple}: GRU model where the input is a concatenation of the measurement vector, which variables are missing, and how long they have been missing: $x_t \leftarrow [x_t;\delta_t;m_t]$.
	\item \textbf{GRU-D}: GRU model with trainable decay rates \cite{cheRecurrentNeuralNetworks2018}.
	\item \textbf{RITS}: A recurrent imputation model that is an extension of GRU-D where missing values are imputed during training \cite{caoBRITSBidirectionalRecurrent2018}.
\end{itemize}

\subsection{Implementation details}

The number of hidden states in I-RNN model is fixed based on the number of input features since the hidden states directly map to individual features. Therefore, we limit the number of hidden states in the RNN baseline models to also have the same number of hidden states as input features so the number of parameters is comparable across models. RNN models are trained with a mini-batch of 512 samples for up-to 200 epochs. Training continued until the validation area under the receiver operating characteristic curve (AUC) did not increase for 10 epochs. In all cases, we found models converged within 100 epochs. The final model was selected as the one having the highest AUC on the validation dataset. We used the Adam optimizer \cite{kingmaAdamMethodStochastic2017} in PyTorch version 1.4.0 \cite{paszkePyTorchImperativeStyle2019} with the search range of learning rate as $\{0.01, 0.001, 0.0003\}$, gradient clipping $\{1, 10, 20\}$, and the beta coefficients for computing the running averages of the gradient fixed at $(0.9, 0.95)$ where the hyperparameters were explored to give the optimal performance on the validation set. The XGBoost model, was trained for $\{200, 400, 800\}$ rounds with feature interactions (depth $\{2,4,8\}$ trees), subsampling training data at each round ($\{100\%,66\%\}$ of the data per-round), subsampling features per round ($\{100\%,66\%\}$ of all features), and a stopping criteria on the validation AUC to prevent overfitting. We trained all models on five random training and validation splits and report the average and standard deviation of the AUC, positive predictive value (PPV), and specificity (Sp) at the breakeven point, which is the threshold where precision equals recall.

\subsection{Data: Hemodynamic Intervention Prediction}

Hemodynamic instability presents with low blood pressure and requires interventions including, vasopressors, fluids, and packed red blood cells (PRBCs) in the case of blood loss \cite{eshelmanMethodologyEvaluatingPerformance2017}. We predict the onset of a significant hemodynamic intervention by dividing patients into stable and unstable cohorts. The eICU Collaborative Research Database dataset was used for the purposes of training and validating the hemodynamic intervention prediction model presented below \cite{pollardEICUCollaborativeResearch2018}. The full dataset is comprised of 3.3 million patient encounters from 364 hospitals across the United States. To ensure that charting of hemodynamic intervention data (e.g., vasopressors, inotropes) were accurate, we restricted our analysis to patients admitted to select hospitals with reliable infusion and ventilation charting data. Specifically, we included ward-years that charted $\geq$ 7 infusion drug entries per patient per day. Included patients with $\geq$0.75 ventilation and airway records per patient per day in the patient care plan, and either $\geq$10 entries per patient per day in respiratory charting tables in eICU database. This filtering step reduced the initial dataset size to 1.4 million patient encounters from 54 hospitals. We selected patients $\ge18$ years old and who did not have a do-not-resuscitate (DNR) order.

Stable patients did not receive vasopressors and large volumes of fluids during their ICU stay. Unstable patients received at least one vasopressor during the ICU stay. These patients ICU stays were further segmented into unstable and intervention periods. An intervention segment started when any of the strong or weak intervention criteria was satisfied following the procedure described in   \cite{eshelmanMethodologyEvaluatingPerformance2017,conroyDynamicEnsembleApproach2016}. The cohort selection criteria resulted in 32,896 unstable events and 183,420 stable events (prevalence=18\%). A stratified subsample of 20\% of the data were held out and reserved for testing of all algorithms, while the remaining 80\% were used to train all models.

We selected variables that are routinely acquired in the ICU, including vital signs, laboratory measurements, and blood gas measurements. The full set of variables included: age, heart rate, invasive and noninvasive systolic blood pressure, mean blood pressure, temperature, noninvasive shock index (ratio of heart rate/systolic blood pressure), central venous pressure, base excess, WBC, SaO2, AST, bands, basophils, BUN, calcium, ionized calcium, CO2, creatinine, eosinophils, glucose, hematocrit, hemoglobin, lactate, magnesium, PaCO2, potassium, PTT, sodium, bilirubin, FiO2, PIP, and mean airway pressure. We require at--least a heart rate and systolic blood pressure be available for the calculation of a risk score during training and evaluation. We select the 6-hours of data leading up to the intervention event (and a random six hour window from a stable patient).

\subsection{Data: In-Hospital Mortality Prediction}

The PhysioNet Challenge 2012 training set A consists of 4000 multivariate clinical time series records of adult ICU patients with 48 clinical variables measured over the first 48 hours of ICU stay \cite{goldbergerPhysioBankPhysioToolkitPhysioNet2000,silvaPredictingInhospitalMortality2012}. Features included vital signs, laboratory measurements, and general descriptors like age, admission weight, and ICU type. We selected 39 of the 48 clinical variables for modeling including: ALP, ALT, AST, Albumin, BUN, Bilirubin, Cholesterol, Creatinine, DiasABP, FiO2, GCS, Glucose, HCO3, HCT, HR, K, Lactate, MAP, MechVent, Mg, NIDiasABP, NIMAP, NISysABP, Na, PaCO2, PaO2, Platelets, RespRate, SaO2, SysABP, Temp, Urine, WBC, pH, Age, Gender, Height, ICUType, and AdmissionWeight. We applied plausibility filters to remove outliers following the procedure outlined in \cite{johnsonPatientSpecificPredictions2012}. Prevalence of in-hospital mortality was 13.9\%. Sequences were limited to a maximum of 150 time steps with longer sequences truncated by removing earlier observations (signal lengths: $74 \pm 22$ time steps).

\section{Results}

In predicting hemodynamic interventions on the Philips eICU dataset, the time series RNN models outperform XGBoost and logistic regression models trained with hand engineered features summarizing the time series signals (Table \ref{tab:hsi}). I-RNN performs as well (AUC: 0.862) as the full-complexity state-of-the-art RNN models GRU-D (AUC: 0.860) and RITS (0.858). However, there was no difference in model performance across any of the models we evaluated in the PhysioNet Challenge 2012 dataset (Table \ref{tab:physionet}). Although the PhysioNet mortality prediction dataset is convenient and widely used as a benchmark for RNN models \cite{caoBRITSBidirectionalRecurrent2018,cheRecurrentNeuralNetworks2018,zhangModellingEHRTimeseries2019a}, it is likely that RNN models are not useful on this task. The sparse time series observed from the first 48 hours of an ICU stay does not contain enough temporal information about disease progression for the RNN based models to exploit to accurately predict in-hospital mortality. In other words, a lot may happen between the first 2-days in the ICU and the last day in the ICU before a patient is discharged or expires so simple summary statistics are as good as a time series model with 2-days of data. The hemodynamic intervention prediction dataset is a better use case for RNNs because hemodynamic intervention decisions at the bedside are largely driven by trends in blood pressure (decreasing), heart rate (increasing), and laboratory measurements (e.g. decreasing hematocrit or hemoglobin levels can indicate blood loss and lead to an intervention with blood transfusion). This is clearly reflected in the model performance where RNN based models significantly outperform non-RNN baselines on the intervention prediction task. In the following sections we describe in-detail how the I-RNN model transparently explains feature importances in the hemodynamic intervention prediction problem.

\begin{table*}[t]
  \centering 
  \caption{Model performance on eICU dataset predicting hemodynamic interventions. XGBoost and Logistic regression use hand engineered features summarizing time series over the 6-hour observation window. RNN models use the raw time series data. Positive predictive value and specificity are reported at the breakeven point where precision equals recall. Prevalence=14\%.}
  \begin{tabular}{|c|c|l|r|r|r|}\hline
    \multicolumn{6}{|c|}{Hemodynamic Interventions} \\ \hline
    Intelligibility & Accuracy & Model & AUC & PPV & Sp \\ \hline
    +++ & + & Logistic & 0.797 (0.001) & 0.473 (0.005) & 0.928 (0.001) \\
    ++ & ++ & XGBoost & 0.848 (0.002) & 0.531 (0.004) & 0.936 (0.001) \\
    + & ++ & GRU-Forward & 0.835 (0.001) & 0.500 (0.007) & 0.932 (0.002) \\
    + & +++ & GRU-Simple & 0.862 (0.001) & 0.544 (0.004) & 0.938 (0.001) \\
    + & +++ & GRU-D & 0.860 (0.003) & 0.546 (0.006) & 0.939 (0.002) \\
    + & +++ & RITS & 0.858 (0.003) & 0.547 (0.007) & 0.939 (0.002) \\
    ++ & +++ & I-RNN & 0.862 (0.003) & 0.537 (0.007) & 0.937 (0.001) \\ \hline
  \end{tabular}
  \label{tab:hsi} 
\end{table*}

\begin{table*}[t]
  \centering 
  \caption{Model performance on PhysioNet Challenge 2012 dataset}
  \begin{tabular}{|l|r|r|r|}\hline
    \multicolumn{4}{|c|}{In-Hospital Mortality} \\ \hline
    Model & AUC & PPV & Sp \\ \hline
    Logistic & 0.834 (0.009) & 0.467 (0.009) & 0.914 (0.010) \\
    XGBoost & 0.844 (0.011) & 0.492 (0.041) & 0.921 (0.009) \\
    GRU-Forward & 0.845 (0.009) & 0.475 (0.016) & 0.916 (0.011) \\
    GRU-Simple & 0.855 (0.014) & 0.470 (0.033) & 0.907 (0.010) \\
    GRU-D & 0.849 (0.004) & 0.468 (0.039) & 0.914 (0.009) \\
    RITS & 0.843 (0.003) & 0.467 (0.017) & 0.916 (0.009) \\
    I-RNN & 0.846 (0.005) & 0.466 (0.030) & 0.913 (0.007) \\
    \hline
  \end{tabular}
  \label{tab:physionet}
\end{table*}

\begin{figure}[t]
  \centering 
  \caption{Individual feature risk curves (red) from the I-RNN model. Feature distributions for stable (blue) and unstable (orange) patients are shown to illustrate that the hidden state reflects the underlying risk. The risk curve is derived by averaging the feature-specific hidden state activations from the I-RNN cell. The risk curve in red is the LOWESS smoothed estimate.}
  \includegraphics[width=\columnwidth]{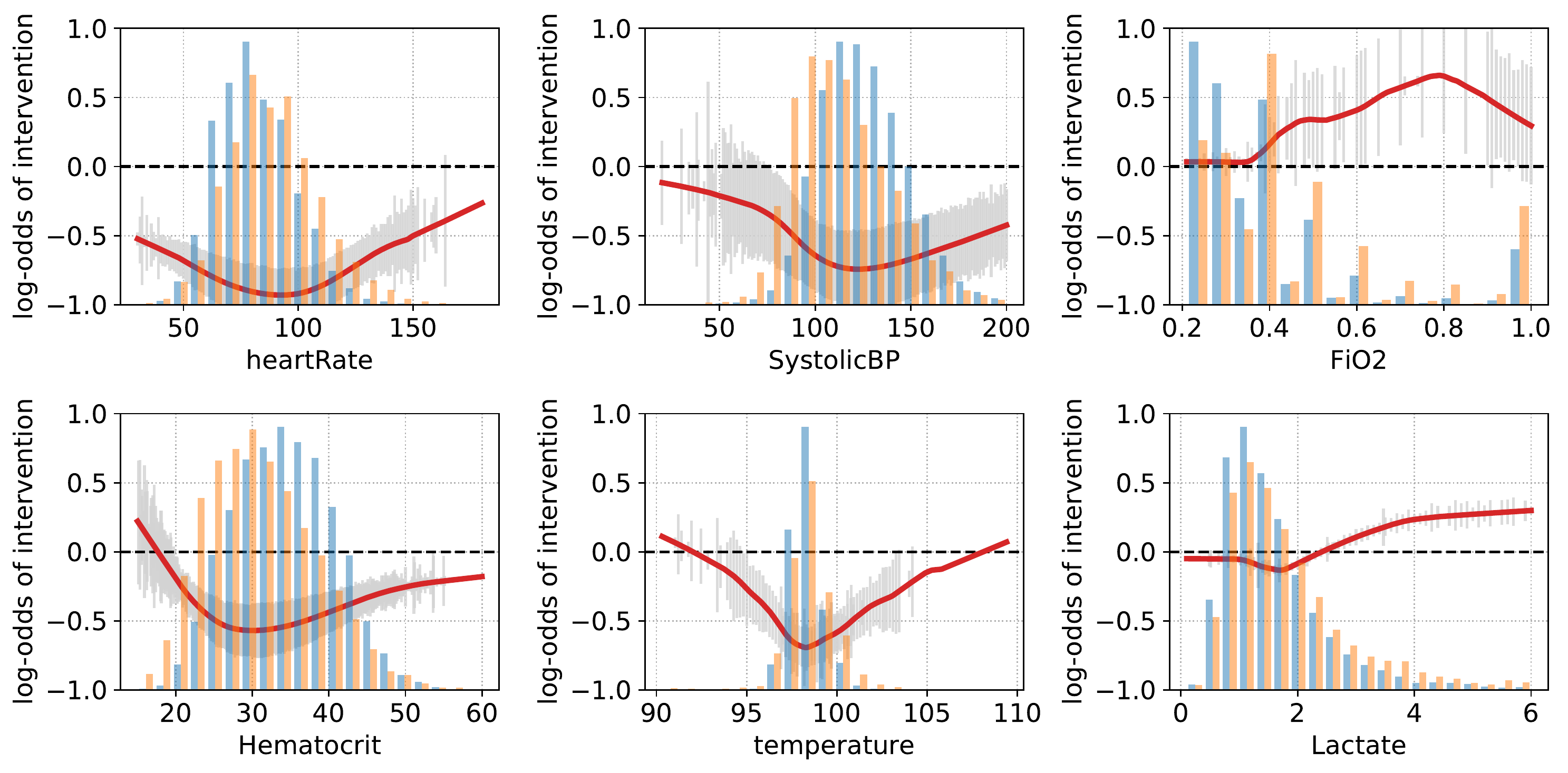}
  \label{fig:curves} 
\end{figure}

Global and local feature importances can be directly visualized from the output of the I-RNN model, an important distinction between I-RNN and the full featured RNN alternatives like GRU-D and RITS. Figure \ref{fig:curves} shows individual risk curves (red) from the hemodynamic intervention model and the underlying feature distributions for the stable (blue) and unstable patients (orange). High heart rate and low systolic blood pressure increases the risk of hemodynamic intervention because these reflect signs of shock. Low hematocrit, possibly as a result of blood loss, increases risk of hemodynamic interventions. Patients with low hematocrit may need blood transfusions, which was one of the hemodynamic interventions we are trying to predict. Similarly, low and high temperatures both increase risk. High lactate levels increase risk of hemodynamic interventions and indicates that lack of oxygenation is a risk for hemodynamic instability.

\begin{figure}[ht]
  \centering 
  \caption{Global feature importances measured by the average impact of features on the model output magnitude. Left: Average hidden states for each variable in the RNN time series model. Right: Average Shapley Additive Explanations (SHAP) values from an XGBoost model.}
  \includegraphics[width=\columnwidth]{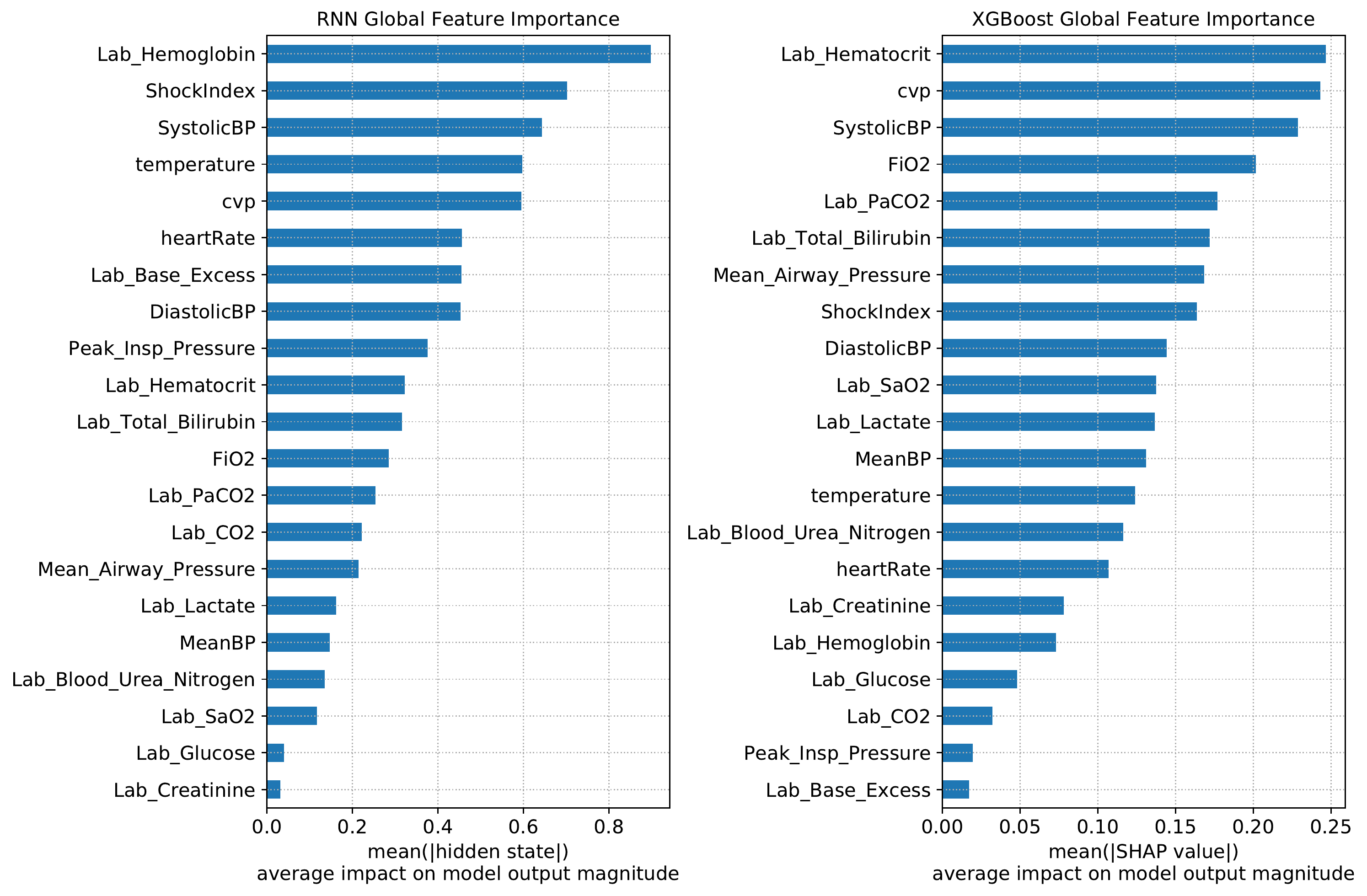}
  \label{fig:global} 
\end{figure}

The global feature importance rankings in Figure \ref{fig:global} are especially interesting since it captures the impact of each feature on the model output. I-RNN learns hemoglobin (highly correlated with hematocrit) and blood pressures are top predictors of hemodynamic instability, similar to the XGBoost model. This further validates the case that the I-RNN can be used to infer associations between input variables and the outcome. We expect the ranking of global feature importances between the time series model and the decision tree model to be similar but not exactly the same. The main reason we expect the rankings to differ is because the XGBoost model does not weight the features by their age. In contrast, the I-RNN model decays the contribution of older variables.

\begin{figure}[t]
  \centering 
  \caption{Example patient case showing the hemodynamic intervention risk score over the 6-hours leading up to intervention along with relevant feature values and their univariate risk scores. The final risk score is a sum of the univariate risks.}
  \includegraphics[width=\columnwidth]{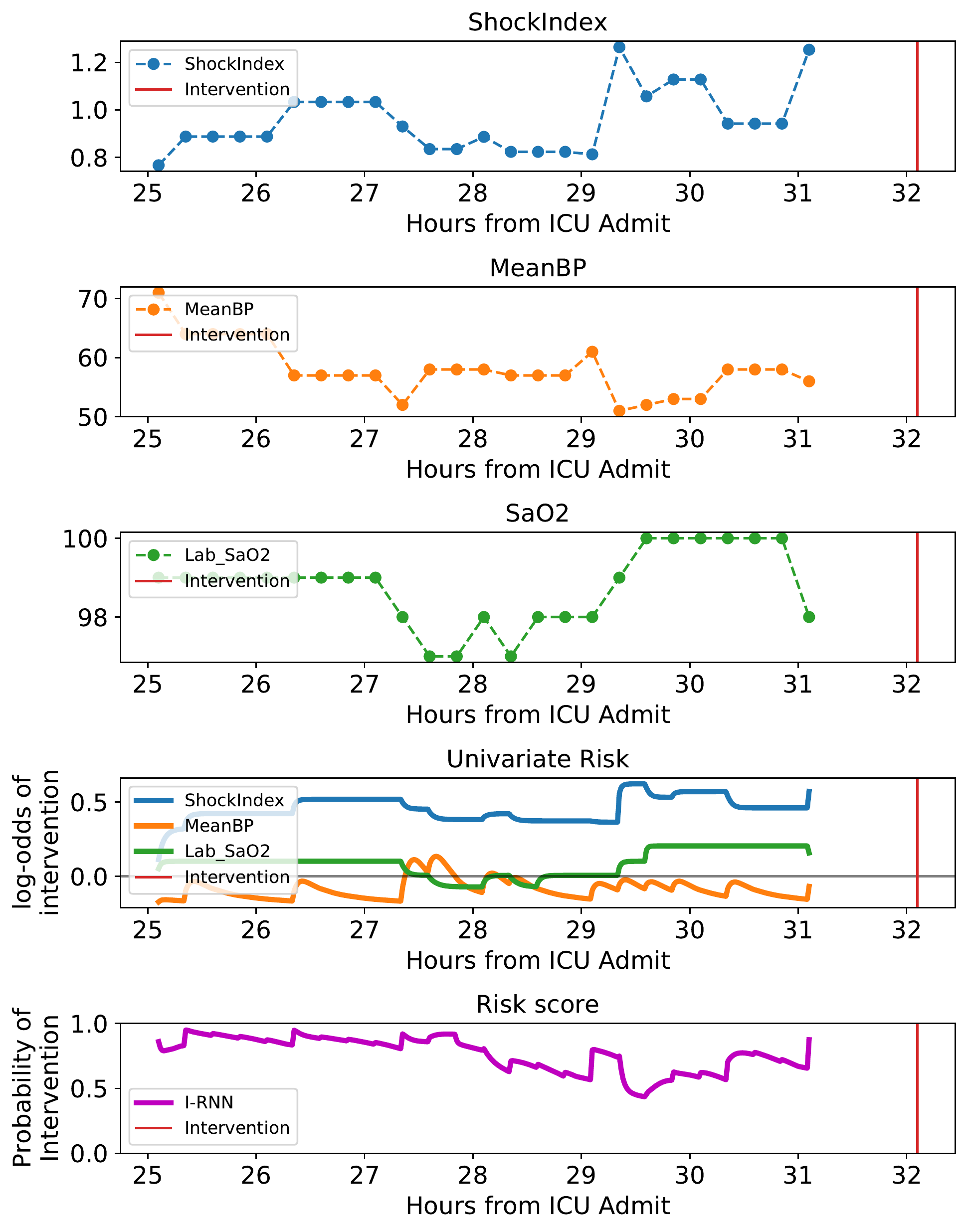}
  \label{fig:example}
\end{figure}

\begin{figure}[t]
  \centering 
  \caption{Decay rates ($\gamma$) as a function of the elapsed time (in hours) between observations ($\delta$). Heart rate, for example, has a fast decay which means only recent observations of heart rate are used by the model, while the contribution of older heart rate measurements are attenuated.}
  \includegraphics[width=\columnwidth]{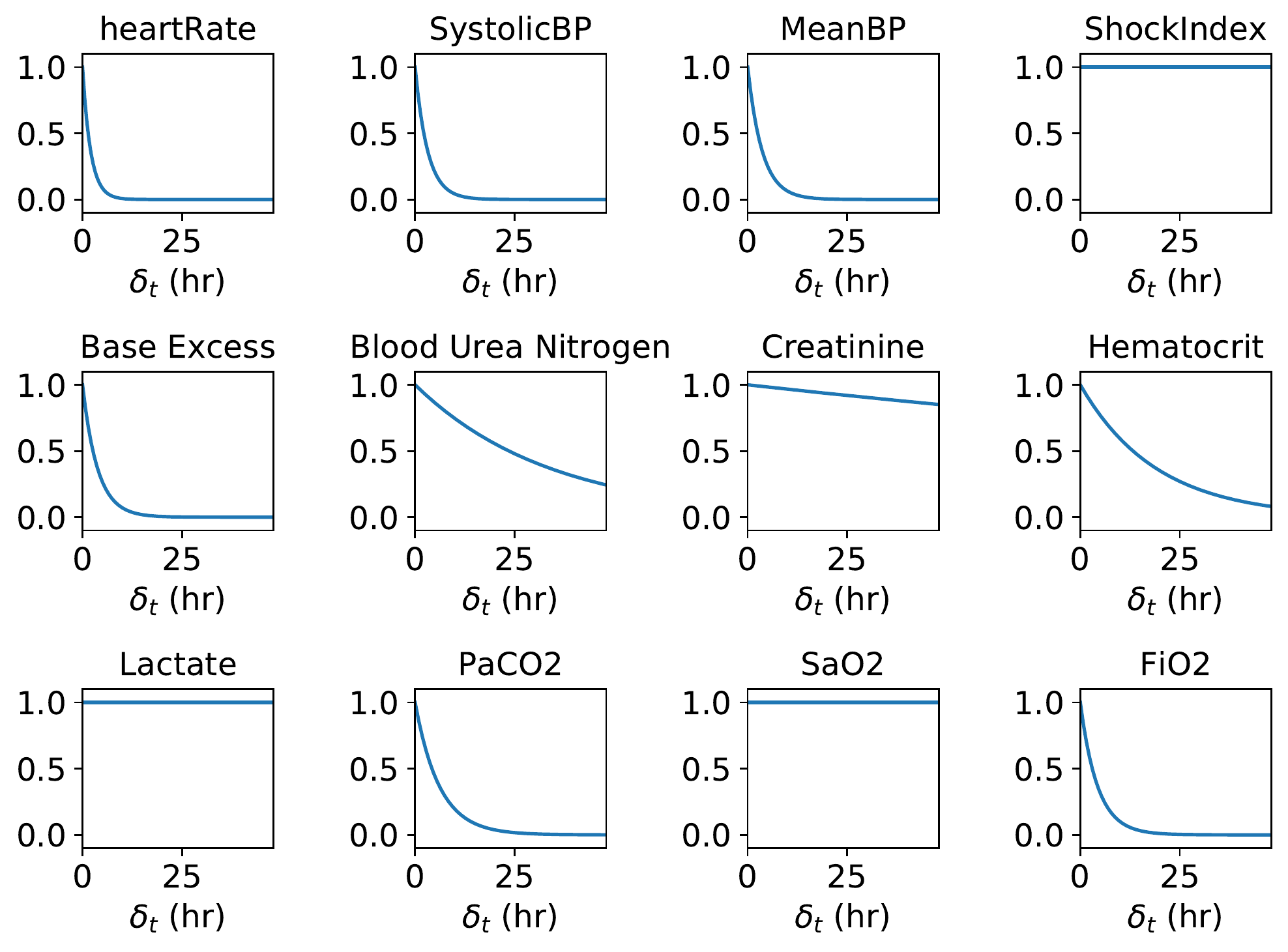}
  \label{fig:decay}
\end{figure}

Figure \ref{fig:example} shows a patient case of the 6-hours leading up to a hemodynamic intervention. The patient has elevated shock index and low mean blood pressure which increases the risk of hemodynamic intervention. The univariate risks for each feature can be visualized as a continuous value. We see that shock index has a slow decay rate so the univariate risk for shock index (blue) does not decrease as much as mean blood pressure (orange), which has a fast decay rate, in-between observations. Decay rates can be visualized as in Figure \ref{fig:decay}, which shows that the contribution from recent observations of heart rate and blood pressures are weighted more than observations made more than 12 hours prior ($\delta_t$ is the age of the current observation).

\begin{figure}[t]
  \centering 
  \caption{Comparison of the univariate risk for shock index learned by I-RNN (magenta) and hidden state activations for GRU-D (green) in the time leading up to the hemodynamic intervention (red). As shock index increases (increasing risk of an intervention), the univariate risk from I-RNN increases along with it. GRU-D is less interpretable because it utilizes missingness information and models interactions between hidden states so it is challenging to map a hidden state to a specific feature. We therefore show the activations of the top four hidden states that are most correlated with shock index.}
  \includegraphics[width=\columnwidth]{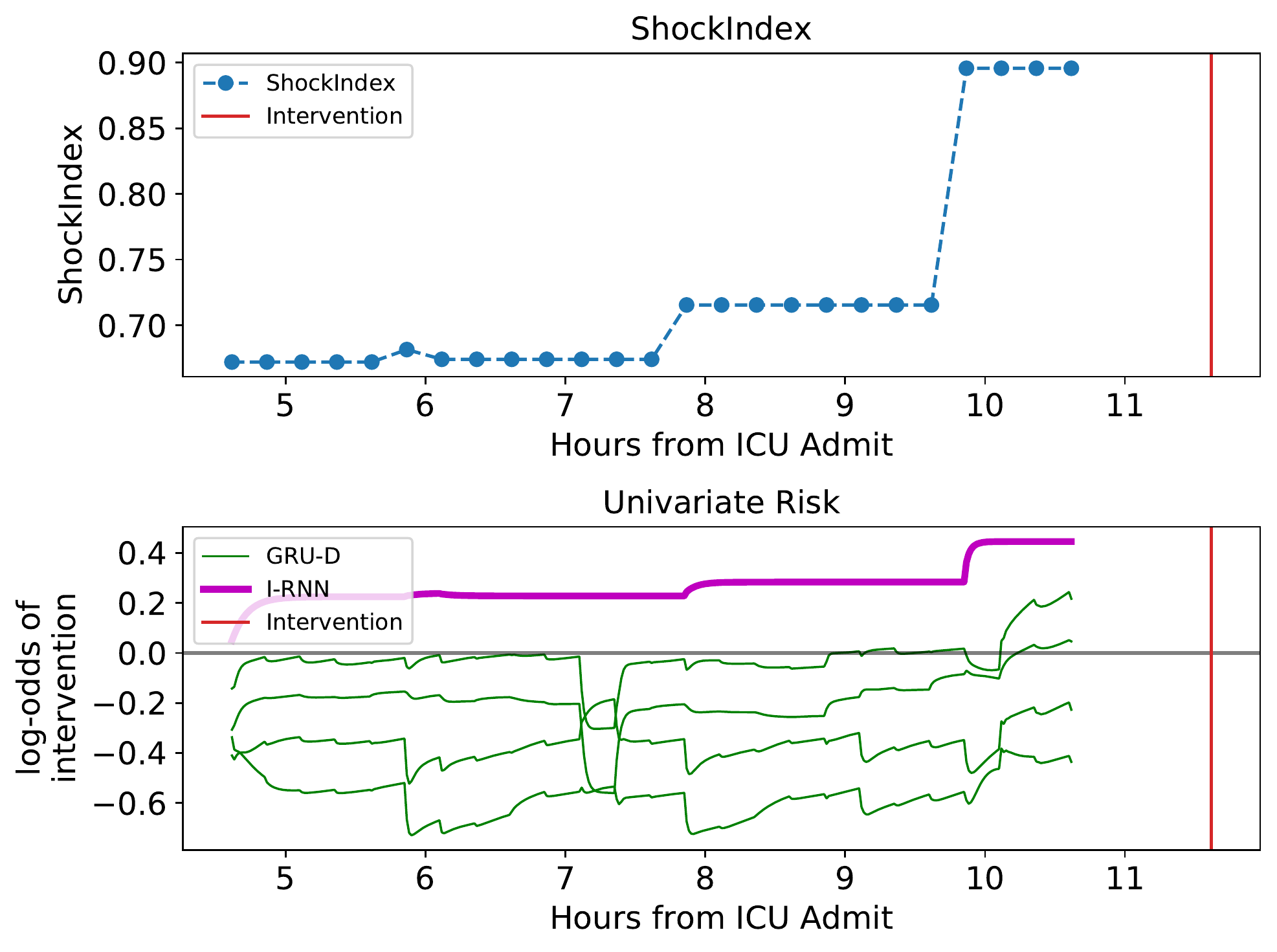}
  \label{fig:uv}
\end{figure}

\begin{figure}[ht]
  \centering 
  \caption{Cross-correlation (green=$+1$, purple=$-1$) between features and hidden states. GRU-D (right) hidden states have interactions with all features but I-RNN (left) is mostly diagonal because of restrictions on interactions.}
  \includegraphics[width=\columnwidth]{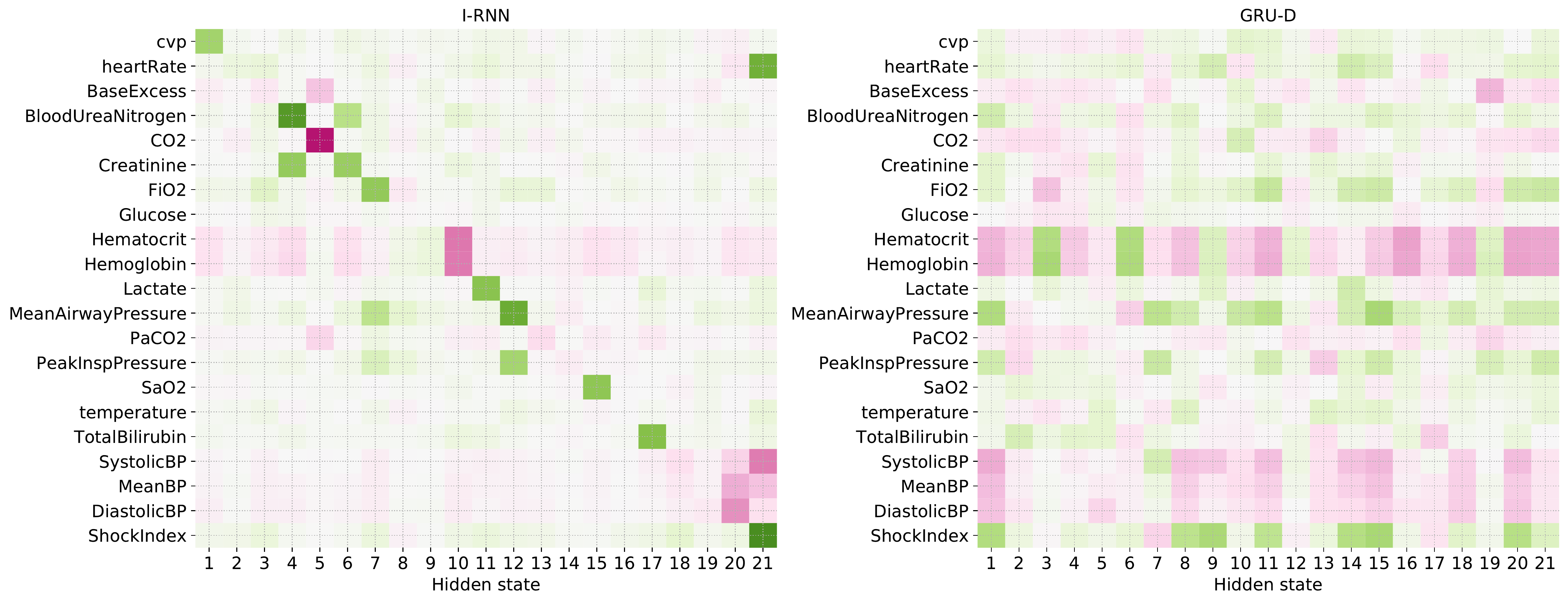}
  \label{fig:corr} 
\end{figure}

Compared to full-complexity models like GRU-D, I-RNN hidden states can be mapped directly to an input feature. Figure \ref{fig:uv} shows that the hidden state encoding shock index increases with increasing shock index, however, hidden states of the GRU-D model don't map to any one feature because hidden states of GRU-D encodes interactions and it uses missing variable indicators as a model input. Figure \ref{fig:corr} shows the cross-correlation (range $[-1, 1]$) between input features and hidden states of the I-RNN and GRU-D models. The restricted feature interactions cause I-RNN activations to correlate strongly with a single feature (diagonal) but hidden states of GRU-D have interactions with all features (off-diagonal).

\section{Discussion}

We achieve an interpretable and accurate time series model by carefully considering the characteristics of multivariate clinical time series signals and by restricting feature interactions for an additive time series model. Our results challenge the concept that there is a tradeoff between accuracy and interpretability.

Table \ref{tab:hsi} summarizes the differences between models of different complexities, borrowing the representation from \cite{louIntelligibleModelsClassification2012}. A linear model like logistic regression is the most intelligible because model coefficients are interpretable as log-odds ratios but are often the least accurate. GRU-Simple, GRU-D, and RITS all utilize informative missingness using the missing variable indicators as model inputs and have interactions, which leads to a less interpretable but highly accurate model. Decision trees, like XGBoost, can model non-linear univariate risk functions and feature interactions which makes them more accurate than linear models. However, trees with depth $>2$ model deep interactions that are difficult to visualize and interpret \cite{louIntelligibleModelsClassification2012}. Post-hoc methods like SHAP and LIME are required to visualize the feature importances from a deep decision tree \cite{lundbergConsistentIndividualizedFeature2019,ribeiroWhyShouldTrust2016}. I-RNN, however, is an additive model that does not use interactions like XGBoost and the GRU models. The result is a high accuracy and interpretable model because the univariate hidden state activations can be mapped back to individual input features.

RNNs are awkward to fit for irregularly sampled time series signals and are better suited for regularly spaced data with few missing samples. A number of approaches have emerged that use the elapsed time between observations and the missingness patterns as additional channels of information. Encoding information about missing variables, however, is problematic in clinical decision support where we are interested in modeling the physiological state from vital signs and laboratory measurements but not necessarily the specific care patterns captured by observation frequency and feature availability. Patterns in missing variables capture clinical practice and clinician concern, which can change between institutions \cite{futomaMythGeneralisabilityClinical2020,reynaEarlyPredictionSepsis2020}. Explicitly modeling care patterns in the form of missing variable indicators are likely to have poor generalization to new care settings with a different care pattern than the training set. Clinician concern reflects the knowledge that the clinician already has, like the patient is deteriorating and laboratory measurements should be taken more frequently. Ideally, we want models that learn physiological risk factors and not necessarily the degree of clinical vigilance. Despite these concerns, RNN models are widely trained to exploit informative missingness patterns \cite{caoBRITSBidirectionalRecurrent2018,cheRecurrentNeuralNetworks2018}, which typically leads to improved model performance but comes at the cost of injecting clinician concern into a model and fitting to institution-specific clinical practice. An important distinction for I-RNN, is that I-RNN does not utilize missingness information as a model input.

Finally, black box models where the function is too complicated for a human to understand can lead to a lack of accountability and to potentially severe consequences like being denied parole \cite{wexlerOpinionWhenComputer2017} or access to medical care \cite{obermeyerDissectingRacialBias2019}. Inherently interpretable models can ensure safety and trust in the models predictions. A transparent model, like I-RNN, is particularly relevant in the healthcare domain where models can learn incorrect associations between input features and outcomes that have to be manually corrected (see pneumonia case study in \cite{caruanaIntelligibleModelsHealthCare2015}). Decomposable additive models where every feature has a coefficient that correlates to the prediction target, like I-RNN, makes it easy to diagnose counterintuitive relationships. For example, we can modify the sign of the output weight ($W_o^d$) if the model learns the opposite association between a feature and the outcome then expected by a clinician. Black box machine learning models that attempt to explain predictions (e.g. using attention-mechanism or post-hoc methods) rather than understanding the underlying relationships between observations and outcomes is likely to lead to a complicated decision path and may not provide enough detail into what the black box is doing \cite{rudinStopExplainingBlack2019}. In contrast, transparent additive models, like I-RNN, can be validated by domain experts simply by looking at the feature coefficients and risk curves.


\bibliographystyle{ACM-Reference-Format}
\bibliography{irnn-rahman.bib}


\begin{thebibliography}{38}


\ifx \showCODEN    \undefined \def \showCODEN     #1{\unskip}     \fi
\ifx \showDOI      \undefined \def \showDOI       #1{#1}\fi
\ifx \showISBNx    \undefined \def \showISBNx     #1{\unskip}     \fi
\ifx \showISBNxiii \undefined \def \showISBNxiii  #1{\unskip}     \fi
\ifx \showISSN     \undefined \def \showISSN      #1{\unskip}     \fi
\ifx \showLCCN     \undefined \def \showLCCN      #1{\unskip}     \fi
\ifx \shownote     \undefined \def \shownote      #1{#1}          \fi
\ifx \showarticletitle \undefined \def \showarticletitle #1{#1}   \fi
\ifx \showURL      \undefined \def \showURL       {\relax}        \fi
\providecommand\bibfield[2]{#2}
\providecommand\bibinfo[2]{#2}
\providecommand\natexlab[1]{#1}
\providecommand\showeprint[2][]{arXiv:#2}

\bibitem[\protect\citeauthoryear{Cao, Wang, Li, Zhou, Li, and Li}{Cao
  et~al\mbox{.}}{[n.d.]}]%
        {caoBRITSBidirectionalRecurrent2018}
\bibfield{author}{\bibinfo{person}{Wei Cao}, \bibinfo{person}{Dong Wang},
  \bibinfo{person}{Jian Li}, \bibinfo{person}{Hao Zhou}, \bibinfo{person}{Lei
  Li}, {and} \bibinfo{person}{Yitan Li}.} \bibinfo{year}{[n.d.]}\natexlab{}.
\newblock \showarticletitle{{{BRITS}}: {{Bidirectional Recurrent Imputation}}
  for {{Time Series}}}.
\newblock  (\bibinfo{year}{[n.\,d.]}), \bibinfo{pages}{11}.
\newblock


\bibitem[\protect\citeauthoryear{Caruana, Lou, Gehrke, Koch, Sturm, and
  Elhadad}{Caruana et~al\mbox{.}}{[n.d.]}]%
        {caruanaIntelligibleModelsHealthCare2015}
\bibfield{author}{\bibinfo{person}{Rich Caruana}, \bibinfo{person}{Yin Lou},
  \bibinfo{person}{Johannes Gehrke}, \bibinfo{person}{Paul Koch},
  \bibinfo{person}{Marc Sturm}, {and} \bibinfo{person}{Noemie Elhadad}.}
  \bibinfo{year}{[n.d.]}\natexlab{}.
\newblock \showarticletitle{Intelligible {{Models}} for {{HealthCare}}:
  {{Predicting Pneumonia Risk}} and {{Hospital}} 30-Day {{Readmission}}}. In
  \bibinfo{booktitle}{\emph{Proceedings of the 21th {{ACM SIGKDD International
  Conference}} on {{Knowledge Discovery}} and {{Data Mining}} - {{KDD}} '15}}
  ({Sydney, NSW, Australia}, 2015). \bibinfo{publisher}{{ACM Press}},
  \bibinfo{pages}{1721--1730}.
\newblock
\showISBNx{978-1-4503-3664-2}
\urldef\tempurl%
\url{https://doi.org/10.1145/2783258.2788613}
\showDOI{\tempurl}


\bibitem[\protect\citeauthoryear{Chakraborty, Tomsett, Raghavendra, Harborne,
  Alzantot, Cerutti, Srivastava, Preece, Julier, Rao, Kelley, Braines, Sensoy,
  Willis, and Gurram}{Chakraborty et~al\mbox{.}}{[n.d.]}]%
        {chakrabortyInterpretabilityDeepLearning2017}
\bibfield{author}{\bibinfo{person}{S. Chakraborty}, \bibinfo{person}{R.
  Tomsett}, \bibinfo{person}{R. Raghavendra}, \bibinfo{person}{D. Harborne},
  \bibinfo{person}{M. Alzantot}, \bibinfo{person}{F. Cerutti},
  \bibinfo{person}{M. Srivastava}, \bibinfo{person}{A. Preece},
  \bibinfo{person}{S. Julier}, \bibinfo{person}{R.~M. Rao},
  \bibinfo{person}{T.~D. Kelley}, \bibinfo{person}{D. Braines},
  \bibinfo{person}{M. Sensoy}, \bibinfo{person}{C.~J. Willis}, {and}
  \bibinfo{person}{P. Gurram}.} \bibinfo{year}{[n.d.]}\natexlab{}.
\newblock \showarticletitle{Interpretability of Deep Learning Models: {{A}}
  Survey of Results}. In \bibinfo{booktitle}{\emph{2017 {{IEEE SmartWorld}},
  {{Ubiquitous Intelligence Computing}}, {{Advanced Trusted Computed}},
  {{Scalable Computing Communications}}, {{Cloud Big Data Computing}},
  {{Internet}} of {{People}} and {{Smart City Innovation}}
  ({{SmartWorld}}/{{SCALCOM}}/{{UIC}}/{{ATC}}/{{CBDCom}}/{{IOP}}/{{SCI}})}}
  (2017-08). \bibinfo{pages}{1--6}.
\newblock
\urldef\tempurl%
\url{https://doi.org/10.1109/UIC-ATC.2017.8397411}
\showDOI{\tempurl}


\bibitem[\protect\citeauthoryear{Che, Purushotham, Cho, Sontag, and Liu}{Che
  et~al\mbox{.}}{[n.d.]a}]%
        {cheRecurrentNeuralNetworks2018}
\bibfield{author}{\bibinfo{person}{Zhengping Che}, \bibinfo{person}{Sanjay
  Purushotham}, \bibinfo{person}{Kyunghyun Cho}, \bibinfo{person}{David
  Sontag}, {and} \bibinfo{person}{Yan Liu}.}
  \bibinfo{year}{[n.d.]}\natexlab{a}.
\newblock \showarticletitle{Recurrent {{Neural Networks}} for {{Multivariate
  Time Series}} with {{Missing Values}}}.
\newblock  \bibinfo{volume}{8}, \bibinfo{number}{1} (\bibinfo{year}{[n.\,d.]}),
  \bibinfo{pages}{6085}.
\newblock
\showISSN{2045-2322}
\urldef\tempurl%
\url{https://doi.org/10.1038/s41598-018-24271-9}
\showDOI{\tempurl}


\bibitem[\protect\citeauthoryear{Che, Purushotham, Khemani, and Liu}{Che
  et~al\mbox{.}}{[n.d.]b}]%
        {cheInterpretableDeepModels2017}
\bibfield{author}{\bibinfo{person}{Zhengping Che}, \bibinfo{person}{Sanjay
  Purushotham}, \bibinfo{person}{Robinder Khemani}, {and} \bibinfo{person}{Yan
  Liu}.} \bibinfo{year}{[n.d.]}\natexlab{b}.
\newblock \showarticletitle{Interpretable {{Deep Models}} for {{ICU Outcome
  Prediction}}}.
\newblock   \bibinfo{volume}{2016} (\bibinfo{year}{[n.\,d.]}),
  \bibinfo{pages}{371--380}.
\newblock
\showISSN{1942-597X}
\showeprint[pmid]{28269832}
\urldef\tempurl%
\url{https://www.ncbi.nlm.nih.gov/pmc/articles/PMC5333206/}
\showURL{%
\tempurl}


\bibitem[\protect\citeauthoryear{Cho, van Merrienboer, Gulcehre, Bahdanau,
  Bougares, Schwenk, and Bengio}{Cho et~al\mbox{.}}{[n.d.]}]%
        {choLearningPhraseRepresentations2014}
\bibfield{author}{\bibinfo{person}{Kyunghyun Cho}, \bibinfo{person}{Bart van
  Merrienboer}, \bibinfo{person}{Caglar Gulcehre}, \bibinfo{person}{Dzmitry
  Bahdanau}, \bibinfo{person}{Fethi Bougares}, \bibinfo{person}{Holger
  Schwenk}, {and} \bibinfo{person}{Yoshua Bengio}.}
  \bibinfo{year}{[n.d.]}\natexlab{}.
\newblock \bibinfo{booktitle}{\emph{Learning {{Phrase Representations}} Using
  {{RNN Encoder}}-{{Decoder}} for {{Statistical Machine Translation}}}}.
\newblock
\showeprint[arxiv]{1406.1078}~[cs, stat]
\urldef\tempurl%
\url{http://arxiv.org/abs/1406.1078}
\showURL{%
\tempurl}


\bibitem[\protect\citeauthoryear{Choi, Bahadori, Kulas, Schuetz, Stewart, and
  Sun}{Choi et~al\mbox{.}}{[n.d.]a}]%
        {choiRETAINInterpretablePredictive2017}
\bibfield{author}{\bibinfo{person}{Edward Choi}, \bibinfo{person}{Mohammad~Taha
  Bahadori}, \bibinfo{person}{Joshua~A. Kulas}, \bibinfo{person}{Andy Schuetz},
  \bibinfo{person}{Walter~F. Stewart}, {and} \bibinfo{person}{Jimeng Sun}.}
  \bibinfo{year}{[n.d.]}\natexlab{a}.
\newblock \bibinfo{booktitle}{\emph{{{RETAIN}}: {{An Interpretable Predictive
  Model}} for {{Healthcare}} Using {{Reverse Time Attention Mechanism}}}}.
\newblock
\showeprint[arxiv]{1608.05745}~[cs]
\urldef\tempurl%
\url{http://arxiv.org/abs/1608.05745}
\showURL{%
\tempurl}


\bibitem[\protect\citeauthoryear{Choi, Bahadori, Schuetz, Stewart, and
  Sun}{Choi et~al\mbox{.}}{[n.d.]b}]%
        {choiDoctorAIPredicting2016}
\bibfield{author}{\bibinfo{person}{Edward Choi}, \bibinfo{person}{Mohammad~Taha
  Bahadori}, \bibinfo{person}{Andy Schuetz}, \bibinfo{person}{Walter~F.
  Stewart}, {and} \bibinfo{person}{Jimeng Sun}.}
  \bibinfo{year}{[n.d.]}\natexlab{b}.
\newblock \showarticletitle{Doctor {{AI}}: {{Predicting Clinical Events}} via
  {{Recurrent Neural Networks}}}.
\newblock   \bibinfo{volume}{56} (\bibinfo{year}{[n.\,d.]}),
  \bibinfo{pages}{301--318}.
\newblock
\showISSN{1938-7288}
\showeprint[pmid]{28286600}
\urldef\tempurl%
\url{https://www.ncbi.nlm.nih.gov/pmc/articles/PMC5341604/}
\showURL{%
\tempurl}


\bibitem[\protect\citeauthoryear{Conroy, Eshelman, Potes, and Xu-Wilson}{Conroy
  et~al\mbox{.}}{[n.d.]}]%
        {conroyDynamicEnsembleApproach2016}
\bibfield{author}{\bibinfo{person}{Bryan Conroy}, \bibinfo{person}{Larry
  Eshelman}, \bibinfo{person}{Cristhian Potes}, {and} \bibinfo{person}{Minnan
  Xu-Wilson}.} \bibinfo{year}{[n.d.]}\natexlab{}.
\newblock \showarticletitle{A Dynamic Ensemble Approach to Robust
  Classification in the Presence of Missing Data}.
\newblock  \bibinfo{volume}{102}, \bibinfo{number}{3}
  (\bibinfo{year}{[n.\,d.]}), \bibinfo{pages}{443--463}.
\newblock
\showISSN{1573-0565}
\urldef\tempurl%
\url{https://doi.org/10.1007/s10994-015-5530-z}
\showDOI{\tempurl}


\bibitem[\protect\citeauthoryear{Eshelman, Xu-Wilson, Flower, Gross, Nielsen,
  Saeed, and Frassica}{Eshelman et~al\mbox{.}}{[n.d.]}]%
        {eshelmanMethodologyEvaluatingPerformance2017}
\bibfield{author}{\bibinfo{person}{Larry~J. Eshelman}, \bibinfo{person}{Minnan
  Xu-Wilson}, \bibinfo{person}{Abigail~A. Flower}, \bibinfo{person}{Brian
  Gross}, \bibinfo{person}{Larry Nielsen}, \bibinfo{person}{Mohammed Saeed},
  {and} \bibinfo{person}{Joseph~J. Frassica}.}
  \bibinfo{year}{[n.d.]}\natexlab{}.
\newblock \bibinfo{title}{A {{Methodology}} for {{Evaluating}} the
  {{Performance}} of {{Alerting}} and {{Detection Algorithms Running}} on
  {{Continuous Patient Data}}}.
\newblock
\newblock
\urldef\tempurl%
\url{https://doi.org/10.1101/182154}
\showDOI{\tempurl}


\bibitem[\protect\citeauthoryear{Futoma, Simons, Panch, Doshi-Velez, and
  Celi}{Futoma et~al\mbox{.}}{[n.d.]}]%
        {futomaMythGeneralisabilityClinical2020}
\bibfield{author}{\bibinfo{person}{Joseph Futoma}, \bibinfo{person}{Morgan
  Simons}, \bibinfo{person}{Trishan Panch}, \bibinfo{person}{Finale
  Doshi-Velez}, {and} \bibinfo{person}{Leo~Anthony Celi}.}
  \bibinfo{year}{[n.d.]}\natexlab{}.
\newblock \showarticletitle{The Myth of Generalisability in Clinical Research
  and Machine Learning in Health Care}.
\newblock  \bibinfo{volume}{2}, \bibinfo{number}{9} (\bibinfo{year}{[n.\,d.]}),
  \bibinfo{pages}{e489--e492}.
\newblock
\showISSN{25897500}
\urldef\tempurl%
\url{https://doi.org/10.1016/S2589-7500(20)30186-2}
\showDOI{\tempurl}


\bibitem[\protect\citeauthoryear{Goldberger, Amaral, Glass, Hausdorff, Ivanov,
  Mark, Mietus, Moody, Peng, and Stanley}{Goldberger et~al\mbox{.}}{[n.d.]}]%
        {goldbergerPhysioBankPhysioToolkitPhysioNet2000}
\bibfield{author}{\bibinfo{person}{Ary~L. Goldberger}, \bibinfo{person}{Luis~AN
  Amaral}, \bibinfo{person}{Leon Glass}, \bibinfo{person}{Jeffrey~M.
  Hausdorff}, \bibinfo{person}{Plamen~Ch Ivanov}, \bibinfo{person}{Roger~G.
  Mark}, \bibinfo{person}{Joseph~E. Mietus}, \bibinfo{person}{George~B. Moody},
  \bibinfo{person}{Chung-Kang Peng}, {and} \bibinfo{person}{H.~Eugene
  Stanley}.} \bibinfo{year}{[n.d.]}\natexlab{}.
\newblock \showarticletitle{{{PhysioBank}}, {{PhysioToolkit}}, and
  {{PhysioNet}}: Components of a New Research Resource for Complex Physiologic
  Signals}.
\newblock  \bibinfo{volume}{101}, \bibinfo{number}{23}
  (\bibinfo{year}{[n.\,d.]}), \bibinfo{pages}{e215--e220}.
\newblock
\showISBNx{0009-7322}


\bibitem[\protect\citeauthoryear{Guo, Lin, and Antulov-Fantulin}{Guo
  et~al\mbox{.}}{[n.d.]}]%
        {guoExploringInterpretableLSTM2019}
\bibfield{author}{\bibinfo{person}{Tian Guo}, \bibinfo{person}{Tao Lin}, {and}
  \bibinfo{person}{Nino Antulov-Fantulin}.} \bibinfo{year}{[n.d.]}\natexlab{}.
\newblock \bibinfo{booktitle}{\emph{Exploring {{Interpretable LSTM Neural
  Networks}} over {{Multi}}-{{Variable Data}}}}.
\newblock
\showeprint[arxiv]{1905.12034}~[cs, stat]
\urldef\tempurl%
\url{http://arxiv.org/abs/1905.12034}
\showURL{%
\tempurl}


\bibitem[\protect\citeauthoryear{Hardt, Rajkomar, Flores, Dai, Howell, Corrado,
  Cui, and Hardt}{Hardt et~al\mbox{.}}{[n.d.]}]%
        {hardtExplainingIncreasePredicted2020a}
\bibfield{author}{\bibinfo{person}{Michaela Hardt}, \bibinfo{person}{Alvin
  Rajkomar}, \bibinfo{person}{Gerardo Flores}, \bibinfo{person}{Andrew Dai},
  \bibinfo{person}{Michael Howell}, \bibinfo{person}{Greg Corrado},
  \bibinfo{person}{Claire Cui}, {and} \bibinfo{person}{Moritz Hardt}.}
  \bibinfo{year}{[n.d.]}\natexlab{}.
\newblock \showarticletitle{Explaining an Increase in Predicted Risk for
  Clinical Alerts}. In \bibinfo{booktitle}{\emph{Proceedings of the {{ACM
  Conference}} on {{Health}}, {{Inference}}, and {{Learning}}}} ({New York, NY,
  USA}, 2020-04-02) \emph{(\bibinfo{series}{{{CHIL}} '20})}.
  \bibinfo{publisher}{{Association for Computing Machinery}},
  \bibinfo{pages}{80--89}.
\newblock
\showISBNx{978-1-4503-7046-2}
\urldef\tempurl%
\url{https://doi.org/10.1145/3368555.3384460}
\showDOI{\tempurl}


\bibitem[\protect\citeauthoryear{Harutyunyan, Khachatrian, Kale, Steeg, and
  Galstyan}{Harutyunyan et~al\mbox{.}}{[n.d.]}]%
        {harutyunyanMultitaskLearningBenchmarking2019a}
\bibfield{author}{\bibinfo{person}{Hrayr Harutyunyan}, \bibinfo{person}{Hrant
  Khachatrian}, \bibinfo{person}{David~C. Kale}, \bibinfo{person}{Greg~Ver
  Steeg}, {and} \bibinfo{person}{Aram Galstyan}.}
  \bibinfo{year}{[n.d.]}\natexlab{}.
\newblock \showarticletitle{Multitask Learning and Benchmarking with Clinical
  Time Series Data}.
\newblock  \bibinfo{volume}{6}, \bibinfo{number}{1} (\bibinfo{year}{[n.\,d.]}),
  \bibinfo{pages}{1--18}.
\newblock
Issue 1.
\showISSN{2052-4463}
\urldef\tempurl%
\url{https://doi.org/10.1038/s41597-019-0103-9}
\showDOI{\tempurl}


\bibitem[\protect\citeauthoryear{Johnson, Dunkley, Mayaud, Tsanas, Kramer, and
  Clifford}{Johnson et~al\mbox{.}}{[n.d.]}]%
        {johnsonPatientSpecificPredictions2012}
\bibfield{author}{\bibinfo{person}{Alistair E.~W. Johnson},
  \bibinfo{person}{Nic Dunkley}, \bibinfo{person}{Louis Mayaud},
  \bibinfo{person}{Athanasios Tsanas}, \bibinfo{person}{Andrew~A Kramer}, {and}
  \bibinfo{person}{Gari~D Clifford}.} \bibinfo{year}{[n.d.]}\natexlab{}.
\newblock \showarticletitle{Patient Specific Predictions in the Intensive Care
  Unit Using a {{Bayesian}} Ensemble}. In \bibinfo{booktitle}{\emph{2012
  {{Computing}} in {{Cardiology}}}} (2012-09). \bibinfo{pages}{249--252}.
\newblock
\showISSN{2325-8853}


\bibitem[\protect\citeauthoryear{Kingma and Ba}{Kingma and Ba}{[n.d.]}]%
        {kingmaAdamMethodStochastic2017}
\bibfield{author}{\bibinfo{person}{Diederik~P. Kingma} {and}
  \bibinfo{person}{Jimmy Ba}.} \bibinfo{year}{[n.d.]}\natexlab{}.
\newblock \bibinfo{booktitle}{\emph{Adam: {{A Method}} for {{Stochastic
  Optimization}}}}.
\newblock
\showeprint[arxiv]{1412.6980}~[cs]
\urldef\tempurl%
\url{http://arxiv.org/abs/1412.6980}
\showURL{%
\tempurl}


\bibitem[\protect\citeauthoryear{Lipton, Kale, Elkan, and Wetzel}{Lipton
  et~al\mbox{.}}{[n.d.]b}]%
        {liptonLearningDiagnoseLSTM2017}
\bibfield{author}{\bibinfo{person}{Zachary~C. Lipton},
  \bibinfo{person}{David~C. Kale}, \bibinfo{person}{Charles Elkan}, {and}
  \bibinfo{person}{Randall Wetzel}.} \bibinfo{year}{[n.d.]}\natexlab{b}.
\newblock \bibinfo{booktitle}{\emph{Learning to {{Diagnose}} with {{LSTM
  Recurrent Neural Networks}}}}.
\newblock
\showeprint[arxiv]{1511.03677}~[cs]
\urldef\tempurl%
\url{http://arxiv.org/abs/1511.03677}
\showURL{%
\tempurl}


\bibitem[\protect\citeauthoryear{Lipton, Kale, and Wetzel}{Lipton
  et~al\mbox{.}}{[n.d.]a}]%
        {liptonModelingMissingData2016}
\bibfield{author}{\bibinfo{person}{Zachary~C. Lipton},
  \bibinfo{person}{David~C. Kale}, {and} \bibinfo{person}{Randall Wetzel}.}
  \bibinfo{year}{[n.d.]}\natexlab{a}.
\newblock \bibinfo{booktitle}{\emph{Modeling {{Missing Data}} in {{Clinical
  Time Series}} with {{RNNs}}}}.
\newblock
\showeprint[arxiv]{1606.04130}~[cs, stat]
\urldef\tempurl%
\url{http://arxiv.org/abs/1606.04130}
\showURL{%
\tempurl}


\bibitem[\protect\citeauthoryear{Lou, Caruana, and Gehrke}{Lou
  et~al\mbox{.}}{[n.d.]}]%
        {louIntelligibleModelsClassification2012}
\bibfield{author}{\bibinfo{person}{Yin Lou}, \bibinfo{person}{Rich Caruana},
  {and} \bibinfo{person}{Johannes Gehrke}.} \bibinfo{year}{[n.d.]}\natexlab{}.
\newblock \showarticletitle{Intelligible Models for Classification and
  Regression}. In \bibinfo{booktitle}{\emph{Proceedings of the 18th {{ACM
  SIGKDD}} International Conference on {{Knowledge}} Discovery and Data Mining
  - {{KDD}} '12}} ({Beijing, China}, 2012). \bibinfo{publisher}{{ACM Press}},
  \bibinfo{pages}{150}.
\newblock
\showISBNx{978-1-4503-1462-6}
\urldef\tempurl%
\url{https://doi.org/10.1145/2339530.2339556}
\showDOI{\tempurl}


\bibitem[\protect\citeauthoryear{Lundberg, Erion, and Lee}{Lundberg
  et~al\mbox{.}}{[n.d.]}]%
        {lundbergConsistentIndividualizedFeature2019}
\bibfield{author}{\bibinfo{person}{Scott~M. Lundberg},
  \bibinfo{person}{Gabriel~G. Erion}, {and} \bibinfo{person}{Su-In Lee}.}
  \bibinfo{year}{[n.d.]}\natexlab{}.
\newblock \bibinfo{booktitle}{\emph{Consistent {{Individualized Feature
  Attribution}} for {{Tree Ensembles}}}}.
\newblock
\showeprint[arxiv]{1802.03888}~[cs, stat]
\urldef\tempurl%
\url{http://arxiv.org/abs/1802.03888}
\showURL{%
\tempurl}


\bibitem[\protect\citeauthoryear{Norgeot, Glicksberg, Trupin, Lituiev,
  Gianfrancesco, Oskotsky, Schmajuk, Yazdany, and Butte}{Norgeot
  et~al\mbox{.}}{[n.d.]}]%
        {norgeotAssessmentDeepLearning2019}
\bibfield{author}{\bibinfo{person}{Beau Norgeot}, \bibinfo{person}{Benjamin~S.
  Glicksberg}, \bibinfo{person}{Laura Trupin}, \bibinfo{person}{Dmytro
  Lituiev}, \bibinfo{person}{Milena Gianfrancesco}, \bibinfo{person}{Boris
  Oskotsky}, \bibinfo{person}{Gabriela Schmajuk}, \bibinfo{person}{Jinoos
  Yazdany}, {and} \bibinfo{person}{Atul~J. Butte}.}
  \bibinfo{year}{[n.d.]}\natexlab{}.
\newblock \showarticletitle{Assessment of a {{Deep Learning Model Based}} on
  {{Electronic Health Record Data}} to {{Forecast Clinical Outcomes}} in
  {{Patients With Rheumatoid Arthritis}}}.
\newblock  \bibinfo{volume}{2}, \bibinfo{number}{3} (\bibinfo{year}{[n.\,d.]}),
  \bibinfo{pages}{e190606}.
\newblock
\showISSN{2574-3805}
\urldef\tempurl%
\url{https://doi.org/10.1001/jamanetworkopen.2019.0606}
\showDOI{\tempurl}
\showeprint[pmid]{30874779}


\bibitem[\protect\citeauthoryear{Obermeyer, Powers, Vogeli, and
  Mullainathan}{Obermeyer et~al\mbox{.}}{[n.d.]}]%
        {obermeyerDissectingRacialBias2019}
\bibfield{author}{\bibinfo{person}{Ziad Obermeyer}, \bibinfo{person}{Brian
  Powers}, \bibinfo{person}{Christine Vogeli}, {and} \bibinfo{person}{Sendhil
  Mullainathan}.} \bibinfo{year}{[n.d.]}\natexlab{}.
\newblock \showarticletitle{Dissecting Racial Bias in an Algorithm Used to
  Manage the Health of Populations}.
\newblock  \bibinfo{volume}{366}, \bibinfo{number}{6464}
  (\bibinfo{year}{[n.\,d.]}), \bibinfo{pages}{447--453}.
\newblock
\showISSN{0036-8075, 1095-9203}
\urldef\tempurl%
\url{https://doi.org/10.1126/science.aax2342}
\showDOI{\tempurl}
\showeprint[pmid]{31649194}


\bibitem[\protect\citeauthoryear{Paszke, Gross, Massa, Lerer, Bradbury, Chanan,
  Killeen, Lin, Gimelshein, Antiga, Desmaison, Kopf, Yang, DeVito, Raison,
  Tejani, Chilamkurthy, Steiner, Fang, Bai, and Chintala}{Paszke
  et~al\mbox{.}}{[n.d.]}]%
        {paszkePyTorchImperativeStyle2019}
\bibfield{author}{\bibinfo{person}{Adam Paszke}, \bibinfo{person}{Sam Gross},
  \bibinfo{person}{Francisco Massa}, \bibinfo{person}{Adam Lerer},
  \bibinfo{person}{James Bradbury}, \bibinfo{person}{Gregory Chanan},
  \bibinfo{person}{Trevor Killeen}, \bibinfo{person}{Zeming Lin},
  \bibinfo{person}{Natalia Gimelshein}, \bibinfo{person}{Luca Antiga},
  \bibinfo{person}{Alban Desmaison}, \bibinfo{person}{Andreas Kopf},
  \bibinfo{person}{Edward Yang}, \bibinfo{person}{Zachary DeVito},
  \bibinfo{person}{Martin Raison}, \bibinfo{person}{Alykhan Tejani},
  \bibinfo{person}{Sasank Chilamkurthy}, \bibinfo{person}{Benoit Steiner},
  \bibinfo{person}{Lu Fang}, \bibinfo{person}{Junjie Bai}, {and}
  \bibinfo{person}{Soumith Chintala}.} \bibinfo{year}{[n.d.]}\natexlab{}.
\newblock \showarticletitle{{{PyTorch}}: {{An Imperative Style}},
  {{High}}-{{Performance Deep Learning Library}}}.
\newblock In \bibinfo{booktitle}{\emph{Advances in {{Neural Information
  Processing Systems}} 32}}, \bibfield{editor}{\bibinfo{person}{H.~Wallach},
  \bibinfo{person}{H.~Larochelle}, \bibinfo{person}{A.~Beygelzimer},
  \bibinfo{person}{F.~d\textbackslash~textquotesingle Alché-Buc},
  \bibinfo{person}{E.~Fox}, {and} \bibinfo{person}{R.~Garnett}} (Eds.).
  \bibinfo{publisher}{{Curran Associates, Inc.}}, \bibinfo{pages}{8026--8037}.
\newblock
\urldef\tempurl%
\url{http://papers.nips.cc/paper/9015-pytorch-an-imperative-style-high-performance-deep-learning-library.pdf}
\showURL{%
\tempurl}


\bibitem[\protect\citeauthoryear{Pollard, Johnson, Raffa, Celi, Mark, and
  Badawi}{Pollard et~al\mbox{.}}{[n.d.]}]%
        {pollardEICUCollaborativeResearch2018}
\bibfield{author}{\bibinfo{person}{Tom~J. Pollard}, \bibinfo{person}{Alistair
  E.~W. Johnson}, \bibinfo{person}{Jesse~D. Raffa}, \bibinfo{person}{Leo~A.
  Celi}, \bibinfo{person}{Roger~G. Mark}, {and} \bibinfo{person}{Omar Badawi}.}
  \bibinfo{year}{[n.d.]}\natexlab{}.
\newblock \showarticletitle{The {{eICU Collaborative Research Database}}, a
  Freely Available Multi-Center Database for Critical Care Research}.
\newblock  \bibinfo{volume}{5}, \bibinfo{number}{1} (\bibinfo{year}{[n.\,d.]}),
  \bibinfo{pages}{1--13}.
\newblock
Issue 1.
\showISSN{2052-4463}
\urldef\tempurl%
\url{https://doi.org/10.1038/sdata.2018.178}
\showDOI{\tempurl}


\bibitem[\protect\citeauthoryear{Rajkomar, Oren, Chen, Dai, Hajaj, Hardt, Liu,
  Liu, Marcus, Sun, Sundberg, Yee, Zhang, Zhang, Flores, Duggan, Irvine, Le,
  Litsch, Mossin, Tansuwan, Wang, Wexler, Wilson, Ludwig, Volchenboum, Chou,
  Pearson, Madabushi, Shah, Butte, Howell, Cui, Corrado, and Dean}{Rajkomar
  et~al\mbox{.}}{[n.d.]}]%
        {rajkomarScalableAccurateDeep2018}
\bibfield{author}{\bibinfo{person}{Alvin Rajkomar}, \bibinfo{person}{Eyal
  Oren}, \bibinfo{person}{Kai Chen}, \bibinfo{person}{Andrew~M. Dai},
  \bibinfo{person}{Nissan Hajaj}, \bibinfo{person}{Michaela Hardt},
  \bibinfo{person}{Peter~J. Liu}, \bibinfo{person}{Xiaobing Liu},
  \bibinfo{person}{Jake Marcus}, \bibinfo{person}{Mimi Sun},
  \bibinfo{person}{Patrik Sundberg}, \bibinfo{person}{Hector Yee},
  \bibinfo{person}{Kun Zhang}, \bibinfo{person}{Yi Zhang},
  \bibinfo{person}{Gerardo Flores}, \bibinfo{person}{Gavin~E. Duggan},
  \bibinfo{person}{Jamie Irvine}, \bibinfo{person}{Quoc Le},
  \bibinfo{person}{Kurt Litsch}, \bibinfo{person}{Alexander Mossin},
  \bibinfo{person}{Justin Tansuwan}, \bibinfo{person}{De Wang},
  \bibinfo{person}{James Wexler}, \bibinfo{person}{Jimbo Wilson},
  \bibinfo{person}{Dana Ludwig}, \bibinfo{person}{Samuel~L. Volchenboum},
  \bibinfo{person}{Katherine Chou}, \bibinfo{person}{Michael Pearson},
  \bibinfo{person}{Srinivasan Madabushi}, \bibinfo{person}{Nigam~H. Shah},
  \bibinfo{person}{Atul~J. Butte}, \bibinfo{person}{Michael~D. Howell},
  \bibinfo{person}{Claire Cui}, \bibinfo{person}{Greg~S. Corrado}, {and}
  \bibinfo{person}{Jeffrey Dean}.} \bibinfo{year}{[n.d.]}\natexlab{}.
\newblock \showarticletitle{Scalable and Accurate Deep Learning with Electronic
  Health Records}.
\newblock  \bibinfo{volume}{1}, \bibinfo{number}{1} (\bibinfo{year}{[n.\,d.]}),
  \bibinfo{pages}{18}.
\newblock
\showISSN{2398-6352}
\urldef\tempurl%
\url{https://doi.org/10.1038/s41746-018-0029-1}
\showDOI{\tempurl}


\bibitem[\protect\citeauthoryear{Reyna, Josef, Jeter, Shashikumar, Westover,
  Nemati, Clifford, and Sharma}{Reyna et~al\mbox{.}}{[n.d.]}]%
        {reynaEarlyPredictionSepsis2020}
\bibfield{author}{\bibinfo{person}{Matthew~A. Reyna},
  \bibinfo{person}{Christopher~S. Josef}, \bibinfo{person}{Russell Jeter},
  \bibinfo{person}{Supreeth~P. Shashikumar}, \bibinfo{person}{M.~Brandon
  Westover}, \bibinfo{person}{Shamim Nemati}, \bibinfo{person}{Gari~D.
  Clifford}, {and} \bibinfo{person}{Ashish Sharma}.}
  \bibinfo{year}{[n.d.]}\natexlab{}.
\newblock \showarticletitle{Early {{Prediction}} of {{Sepsis From Clinical
  Data}}: {{The PhysioNet}}/{{Computing}} in {{Cardiology Challenge}} 2019}.
\newblock  \bibinfo{volume}{48}, \bibinfo{number}{2}
  (\bibinfo{year}{[n.\,d.]}), \bibinfo{pages}{210--217}.
\newblock
\showISSN{0090-3493}
\urldef\tempurl%
\url{https://doi.org/10.1097/CCM.0000000000004145}
\showDOI{\tempurl}
\showeprint[pmid]{31939789}


\bibitem[\protect\citeauthoryear{Ribeiro, Singh, and Guestrin}{Ribeiro
  et~al\mbox{.}}{[n.d.]}]%
        {ribeiroWhyShouldTrust2016}
\bibfield{author}{\bibinfo{person}{Marco~Tulio Ribeiro},
  \bibinfo{person}{Sameer Singh}, {and} \bibinfo{person}{Carlos Guestrin}.}
  \bibinfo{year}{[n.d.]}\natexlab{}.
\newblock \bibinfo{booktitle}{\emph{"{{Why Should I Trust You}}?":
  {{Explaining}} the {{Predictions}} of {{Any Classifier}}}}.
\newblock
\showeprint[arxiv]{1602.04938}~[cs, stat]
\urldef\tempurl%
\url{http://arxiv.org/abs/1602.04938}
\showURL{%
\tempurl}


\bibitem[\protect\citeauthoryear{Rudin}{Rudin}{[n.d.]}]%
        {rudinStopExplainingBlack2019}
\bibfield{author}{\bibinfo{person}{Cynthia Rudin}.}
  \bibinfo{year}{[n.d.]}\natexlab{}.
\newblock \showarticletitle{Stop Explaining Black Box Machine Learning Models
  for High Stakes Decisions and Use Interpretable Models Instead}.
\newblock  \bibinfo{volume}{1}, \bibinfo{number}{5} (\bibinfo{year}{[n.\,d.]}),
  \bibinfo{pages}{206--215}.
\newblock
\showISSN{2522-5839}
\urldef\tempurl%
\url{https://doi.org/10.1038/s42256-019-0048-x}
\showDOI{\tempurl}


\bibitem[\protect\citeauthoryear{Sherman, Gurm, Balis, Owens, and
  Wiens}{Sherman et~al\mbox{.}}{[n.d.]}]%
        {shermanLeveragingClinicalTimeSeries2018}
\bibfield{author}{\bibinfo{person}{Eli Sherman}, \bibinfo{person}{Hitinder
  Gurm}, \bibinfo{person}{Ulysses Balis}, \bibinfo{person}{Scott Owens}, {and}
  \bibinfo{person}{Jenna Wiens}.} \bibinfo{year}{[n.d.]}\natexlab{}.
\newblock \showarticletitle{Leveraging {{Clinical Time}}-{{Series Data}} for
  {{Prediction}}: {{A Cautionary Tale}}}.
\newblock   \bibinfo{volume}{2017} (\bibinfo{year}{[n.\,d.]}),
  \bibinfo{pages}{1571--1580}.
\newblock
\showISSN{1942-597X}
\showeprint[pmid]{29854227}
\urldef\tempurl%
\url{https://www.ncbi.nlm.nih.gov/pmc/articles/PMC5977714/}
\showURL{%
\tempurl}


\bibitem[\protect\citeauthoryear{Silva, Moody, Scott, Celi, and Mark}{Silva
  et~al\mbox{.}}{[n.d.]}]%
        {silvaPredictingInhospitalMortality2012}
\bibfield{author}{\bibinfo{person}{Ikaro Silva}, \bibinfo{person}{George
  Moody}, \bibinfo{person}{Daniel~J. Scott}, \bibinfo{person}{Leo~A. Celi},
  {and} \bibinfo{person}{Roger~G. Mark}.} \bibinfo{year}{[n.d.]}\natexlab{}.
\newblock \showarticletitle{Predicting In-Hospital Mortality of Icu Patients:
  {{The}} Physionet/Computing in Cardiology Challenge 2012}. In
  \bibinfo{booktitle}{\emph{2012 {{Computing}} in {{Cardiology}}}} (2012).
  \bibinfo{publisher}{{IEEE}}, \bibinfo{pages}{245--248}.
\newblock
\showISBNx{1-4673-2077-3}


\bibitem[\protect\citeauthoryear{Tan, Caruana, Hooker, Koch, and Gordo}{Tan
  et~al\mbox{.}}{[n.d.]}]%
        {tanLearningGlobalAdditive2018}
\bibfield{author}{\bibinfo{person}{Sarah Tan}, \bibinfo{person}{Rich Caruana},
  \bibinfo{person}{Giles Hooker}, \bibinfo{person}{Paul Koch}, {and}
  \bibinfo{person}{Albert Gordo}.} \bibinfo{year}{[n.d.]}\natexlab{}.
\newblock \bibinfo{booktitle}{\emph{Learning {{Global Additive Explanations}}
  for {{Neural Nets Using Model Distillation}}}}.
\newblock
\showeprint[arxiv]{1801.08640}~[cs, stat]
\urldef\tempurl%
\url{http://arxiv.org/abs/1801.08640}
\showURL{%
\tempurl}


\bibitem[\protect\citeauthoryear{Tomašev, Glorot, Rae, Zielinski, Askham,
  Saraiva, Mottram, Meyer, Ravuri, Protsyuk, Connell, Hughes, Karthikesalingam,
  Cornebise, Montgomery, Rees, Laing, Baker, Peterson, Reeves, Hassabis, King,
  Suleyman, Back, Nielson, Ledsam, and Mohamed}{Tomašev
  et~al\mbox{.}}{[n.d.]}]%
        {tomasevClinicallyApplicableApproach2019}
\bibfield{author}{\bibinfo{person}{Nenad Tomašev}, \bibinfo{person}{Xavier
  Glorot}, \bibinfo{person}{Jack~W. Rae}, \bibinfo{person}{Michal Zielinski},
  \bibinfo{person}{Harry Askham}, \bibinfo{person}{Andre Saraiva},
  \bibinfo{person}{Anne Mottram}, \bibinfo{person}{Clemens Meyer},
  \bibinfo{person}{Suman Ravuri}, \bibinfo{person}{Ivan Protsyuk},
  \bibinfo{person}{Alistair Connell}, \bibinfo{person}{Cían~O. Hughes},
  \bibinfo{person}{Alan Karthikesalingam}, \bibinfo{person}{Julien Cornebise},
  \bibinfo{person}{Hugh Montgomery}, \bibinfo{person}{Geraint Rees},
  \bibinfo{person}{Chris Laing}, \bibinfo{person}{Clifton~R. Baker},
  \bibinfo{person}{Kelly Peterson}, \bibinfo{person}{Ruth Reeves},
  \bibinfo{person}{Demis Hassabis}, \bibinfo{person}{Dominic King},
  \bibinfo{person}{Mustafa Suleyman}, \bibinfo{person}{Trevor Back},
  \bibinfo{person}{Christopher Nielson}, \bibinfo{person}{Joseph~R. Ledsam},
  {and} \bibinfo{person}{Shakir Mohamed}.} \bibinfo{year}{[n.d.]}\natexlab{}.
\newblock \showarticletitle{A Clinically Applicable Approach to Continuous
  Prediction of Future Acute Kidney Injury}.
\newblock  \bibinfo{volume}{572}, \bibinfo{number}{7767}
  (\bibinfo{year}{[n.\,d.]}), \bibinfo{pages}{116--119}.
\newblock
\showISSN{0028-0836, 1476-4687}
\urldef\tempurl%
\url{https://doi.org/10.1038/s41586-019-1390-1}
\showDOI{\tempurl}


\bibitem[\protect\citeauthoryear{Wang, Sha, Lakin, Bynum, Bates, Hong, and
  Zhou}{Wang et~al\mbox{.}}{[n.d.]}]%
        {wangDevelopmentValidationDeep2019}
\bibfield{author}{\bibinfo{person}{Liqin Wang}, \bibinfo{person}{Long Sha},
  \bibinfo{person}{Joshua~R. Lakin}, \bibinfo{person}{Julie Bynum},
  \bibinfo{person}{David~W. Bates}, \bibinfo{person}{Pengyu Hong}, {and}
  \bibinfo{person}{Li Zhou}.} \bibinfo{year}{[n.d.]}\natexlab{}.
\newblock \showarticletitle{Development and {{Validation}} of a {{Deep Learning
  Algorithm}} for {{Mortality Prediction}} in {{Selecting Patients With
  Dementia}} for {{Earlier Palliative Care Interventions}}}.
\newblock  \bibinfo{volume}{2}, \bibinfo{number}{7} (\bibinfo{year}{[n.\,d.]}),
  \bibinfo{pages}{e196972--e196972}.
\newblock
\urldef\tempurl%
\url{https://doi.org/10.1001/jamanetworkopen.2019.6972}
\showDOI{\tempurl}


\bibitem[\protect\citeauthoryear{Wexler}{Wexler}{[n.d.]}]%
        {wexlerOpinionWhenComputer2017}
\bibfield{author}{\bibinfo{person}{Rebecca Wexler}.}
  \bibinfo{year}{[n.d.]}\natexlab{}.
\newblock \showarticletitle{Opinion | {{When}} a {{Computer Program Keeps You}}
  in {{Jail}}}.
\newblock  (\bibinfo{year}{[n.\,d.]}).
\newblock
\showISSN{0362-4331}
\urldef\tempurl%
\url{https://www.nytimes.com/2017/06/13/opinion/how-computers-are-harming-criminal-justice.html}
\showURL{%
\tempurl}


\bibitem[\protect\citeauthoryear{Xu, Biswal, Deshpande, Maher, and Sun}{Xu
  et~al\mbox{.}}{[n.d.]}]%
        {xuRAIMRecurrentAttentive2018}
\bibfield{author}{\bibinfo{person}{Yanbo Xu}, \bibinfo{person}{Siddharth
  Biswal}, \bibinfo{person}{Shriprasad~R. Deshpande}, \bibinfo{person}{Kevin~O.
  Maher}, {and} \bibinfo{person}{Jimeng Sun}.}
  \bibinfo{year}{[n.d.]}\natexlab{}.
\newblock \bibinfo{booktitle}{\emph{{{RAIM}}: {{Recurrent Attentive}} and
  {{Intensive Model}} of {{Multimodal Patient Monitoring Data}}}}.
\newblock
\showeprint[arxiv]{1807.08820}~[cs, stat]
\urldef\tempurl%
\url{http://arxiv.org/abs/1807.08820}
\showURL{%
\tempurl}


\bibitem[\protect\citeauthoryear{Zhang, Kowsari, Harrison, Lobo, and
  Barnes}{Zhang et~al\mbox{.}}{[n.d.]a}]%
        {zhangPatient2VecPersonalizedInterpretable2018}
\bibfield{author}{\bibinfo{person}{Jinghe Zhang}, \bibinfo{person}{Kamran
  Kowsari}, \bibinfo{person}{James~H. Harrison}, \bibinfo{person}{Jennifer~M.
  Lobo}, {and} \bibinfo{person}{Laura~E. Barnes}.}
  \bibinfo{year}{[n.d.]}\natexlab{a}.
\newblock \showarticletitle{{{Patient2Vec}}: {{A Personalized Interpretable
  Deep Representation}} of the {{Longitudinal Electronic Health Record}}}.
\newblock   \bibinfo{volume}{6} (\bibinfo{year}{[n.\,d.]}),
  \bibinfo{pages}{65333--65346}.
\newblock
\showISSN{2169-3536}
\urldef\tempurl%
\url{https://doi.org/10.1109/ACCESS.2018.2875677}
\showDOI{\tempurl}
\showeprint[arxiv]{1810.04793}


\bibitem[\protect\citeauthoryear{Zhang, Xue, Flores, Rajkomar, Cui, and
  Dai}{Zhang et~al\mbox{.}}{[n.d.]b}]%
        {zhangModellingEHRTimeseries2019a}
\bibfield{author}{\bibinfo{person}{Kun Zhang}, \bibinfo{person}{Yuan Xue},
  \bibinfo{person}{Gerardo Flores}, \bibinfo{person}{Alvin Rajkomar},
  \bibinfo{person}{Claire Cui}, {and} \bibinfo{person}{Andrew~M. Dai}.}
  \bibinfo{year}{[n.d.]}\natexlab{b}.
\newblock \bibinfo{booktitle}{\emph{Modelling {{EHR}} Timeseries by Restricting
  Feature Interaction}}.
\newblock
\showeprint[arxiv]{1911.06410}~[cs, stat]
\urldef\tempurl%
\url{http://arxiv.org/abs/1911.06410}
\showURL{%
\tempurl}


\end{thebibliography}

\end{document}